\documentclass[10pt,twocolumn,letterpaper]{article}

\usepackage{cvpr}              % To produce the CAMERA-READY version
\definecolor{my_pink}{RGB}{248,206,204}
\definecolor{my_blue}{RGB}{212,225,245}
\definecolor{my_yellow}{RGB}{255,242,204}

\usepackage{arydshln}

\makeatletter
\def\adl@drawiv#1#2#3{%
        \hskip.5\tabcolsep
        \xleaders#3{#2.5\@tempdimb #1{1}#2.5\@tempdimb}%
                #2\z@ plus1fil minus1fil\relax
        \hskip.5\tabcolsep}
\newcommand{\cdashlinelr}[1]{%
  \noalign{\vskip\aboverulesep
           \global\let\@dashdrawstore\adl@draw
           \global\let\adl@draw\adl@drawiv}
  \cdashline{#1}
  \noalign{\global\let\adl@draw\@dashdrawstore
           \vskip\belowrulesep}}
\makeatother

\definecolor{cvprblue}{rgb}{0.21,0.49,0.74}
\usepackage[pagebackref,breaklinks,colorlinks,allcolors=cvprblue]{hyperref}

\usepackage{multirow}
\usepackage{adjustbox}
\usepackage{algorithm}
\usepackage{algpseudocode}
\usepackage[accsupp]{axessibility} 
\newcommand\our{PAR\xspace}
\algrenewcommand\algorithmicrequire{\textbf{Input:}}
\algrenewcommand\algorithmicensure{\textbf{Output:}}
\usepackage[numbers,sort&compress]{natbib}
\setcitestyle{numbers,square,comma}

\newcommand{\blockcomment}[1]{}
\makeatletter
\def\blfootnote{\xdef\@thefnmark{}\@footnotetext}
\makeatother

\def\paperID{6268} % *** Enter the Paper ID here
\def\confName{CVPR}
\def\confYear{2025}

\title{Parallelized Autoregressive Visual Generation}
\author{
  \textbf{Yuqing Wang}\textsuperscript{1} 
  \quad \textbf{Shuhuai Ren}\textsuperscript{3} 
  \quad \textbf{Zhijie Lin}\textsuperscript{2$\dagger$} 
  \quad \textbf{Yujin Han}\textsuperscript{1} \\
  \quad \textbf{Haoyuan Guo}\textsuperscript{2} 
  \quad \textbf{Zhenheng Yang}\textsuperscript{2} 
  \quad \textbf{Difan Zou}\textsuperscript{1} 
  \quad \textbf{Jiashi Feng}\textsuperscript{2}  
  \quad \textbf{Xihui Liu}\textsuperscript{1}\footnotemark[1]  \\ 
\textsuperscript{1}University of Hong Kong  \quad
\textsuperscript{2}ByteDance \quad
\textsuperscript{3}Peking University \\
}

\usepackage[numbers,sort&compress]{natbib}
\setcitestyle{numbers,square,comma}
\usepackage{booktabs}
\usepackage{hhline}
\usepackage{makecell} 
\usepackage{dsfont}

\begin{document}
\maketitle
\blfootnote{$\dagger$Project lead. \quad $*$Corresponding author.}
\begin{abstract}

Autoregressive models have emerged as a powerful approach for visual generation but suffer from slow inference speed due to their sequential token-by-token prediction process.
In this paper, we propose a simple yet effective approach for parallelized autoregressive visual generation that improves generation efficiency while preserving the advantages of autoregressive modeling.
Our key insight is that parallel generation depends on visual token dependencies\textemdash tokens with weak dependencies can be generated in parallel, while strongly dependent adjacent tokens are difficult to generate together, as their independent sampling may lead to inconsistencies.
Based on this observation, we develop a parallel generation strategy that generates distant tokens with weak dependencies in parallel while maintaining sequential generation for strongly dependent local tokens. 
Our approach can be seamlessly integrated into standard autoregressive models without modifying the architecture or tokenizer. 
Experiments on ImageNet and UCF-101 demonstrate that our method achieves a 3.6$\times$ speedup with comparable quality and up to 9.5$\times$ speedup with minimal quality degradation across both image and video generation tasks.
We hope this work will inspire future research in efficient visual generation and unified autoregressive modeling. Project page: \url{https://yuqingwang1029.github.io/PAR-project}. 

\end{abstract}    
\section{Introduction}
\label{sec:intro}

Autoregressive modeling has achieved remarkable success in language modeling~\cite{radford2018improving, radford2019language, gpt3,touvron2023llama,touvron2023llama2}, inspiring its application to visual generation~\cite{van2016conditional, van2016pixel, salimans2017pixelcnn++, parmar2018image, chen2020generative, ramesh2021dalle, esser2020taming,yu2022scaling,tian2024visual, sun2024autoregressive,rq,pang2024randar,chen2024next,ren2025nbp}.
These models show great potential for visual tasks due to their strong scalability and unified modeling capabilities~\cite{henighan2020scaling,team2024chameleon, wang2024emu3}.
% The strong scalability and unified modeling capabilities of autoregressive models make them particularly promising for visual tasks~\cite{henighan2020scaling,team2024chameleon, wang2024emu3}.
Current autoregressive visual generation approaches typically rely on a sequential token-by-token generation paradigm: visual data is first encoded into token sequences using an autoencoder~\cite{vqvae,esser2020taming,yu2023language}, then an autoregressive transformer~\cite{vaswani2017attention} is trained to predict these tokens following a raster scan order~\cite{chen2020generative}. 
However, this strictly sequential generation process leads to a slow generation speed,
severely limiting its practical applications~\citep{liu2024lumina, wang2024emu3}.
In this work, we aim to develop an efficient autoregressive visual generation approach that improves generation speed while maintaining the generation quality.

\begin{figure}[tbp]
  \centering
  \includegraphics[width=\linewidth]{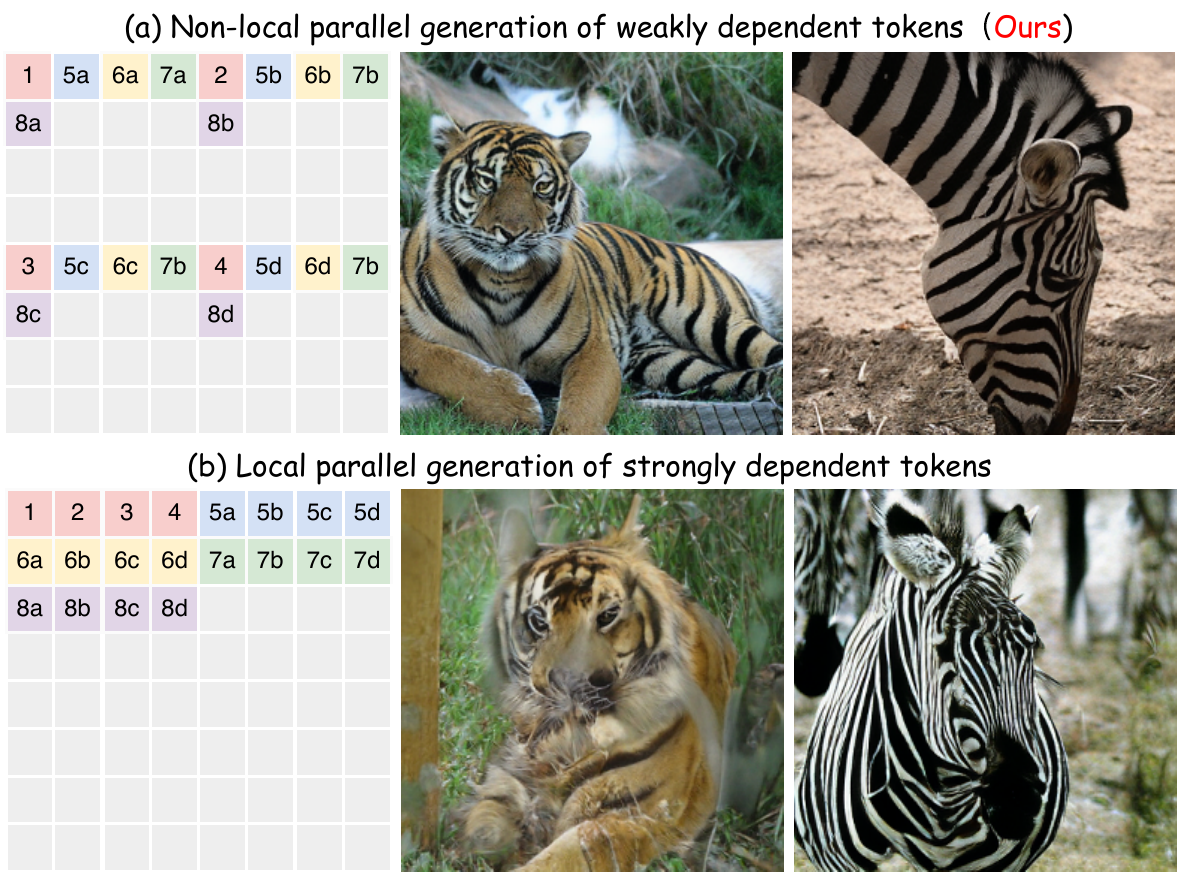}
  \vspace{-16pt}
  \caption{\textbf{Comparison of different parallel generation strategies.} Both strategies generate initial tokens [1,2,3,4] sequentially then generate multiple tokens in parallel per step, following the order [5a-5d] to [6a-6d] to [7a-7d], etc. (a) Our approach generates weakly dependent tokens across non-local regions in parallel, preserving coherent patterns and local details. (b) The naive method generates strongly dependent tokens within local regions simultaneously, while independent sampling for strongly correlated tokens can cause inconsistent generation and disrupted patterns, such as distorted tiger faces and fragmented zebra stripes.}
  % without access to necessary dependent tokens, 
  \label{fig:teaser}
   \vspace{-16pt}
\end{figure}

\begin{figure*}[htbp]
   \centering
   \includegraphics[width=\linewidth]{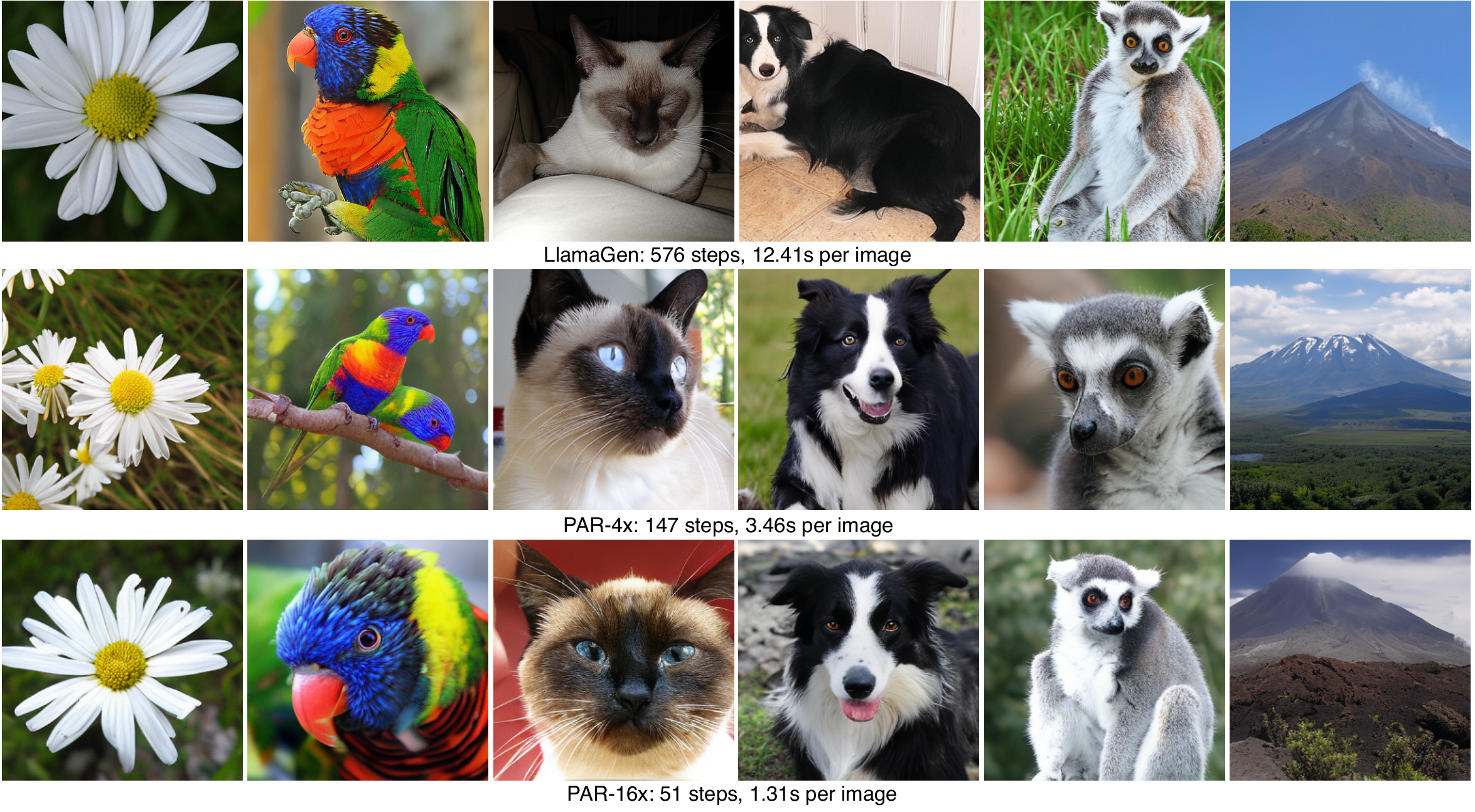}
   \vspace{-22pt}
   \caption{\textbf{Visualization comparison of our parallel generation and traditional autoregressive generation (LlamaGen~\cite{sun2024autoregressive})}. Our approach (PAR) achieves \textbf{3.6-9.5}$\times$ speedup over LlamaGen with comparable quality, reducing the generation time from 12.41s to 3.46s (PAR-4$\times$) and 1.31s (PAR-16$\times$) per image. Time measurements are conducted with a batch size of 1 on a single A100 GPU.
   }
   \vspace{-8pt}
   \label{fig:intro_vis}
\end{figure*}

An intuitive way to improve generation efficiency is to predict multiple tokens in parallel at each step. In language modeling, methods like speculative decoding~\cite{leviathan2023fast,chen2023accelerating,li2024eagle} and Jacobi decoding~\cite{song2021accelerating,kou2024cllms} achieve parallel generation through auxiliary draft models or iterative refinement. In the visual domain, approaches like MaskGIT~\cite{chang2022maskgit} employ non-autoregressive paradigms with masked modeling strategies, while VAR~\cite{tian2024visual} achieves faster speed through next-scale prediction that requires specially designed multi-scale tokenizers and longer token sequences. However, the introduction of additional models and specialized architectures increases model complexity, and may limit the flexibility of autoregressive models as a unified solution across different modalities.

In this work, we ask: \emph{can we achieve parallel visual generation while maintaining the simplicity and flexibility of standard autoregressive models?}  We find that parallel generation is closely tied to token dependencies\textemdash tokens with strong dependencies need sequential generation, while weakly dependent tokens can be generated in parallel. In autoregressive models, each token is generated through sampling (e.g., top-k) to maintain diversity. Parallel generation requires independent sampling of multiple tokens simultaneously, but the joint distribution of highly dependent tokens cannot be factorized for independent sampling, leading to inconsistent predictions, as demonstrated by the distorted local patterns in Fig.~\ref{fig:teaser} (b).
% Such independent sampling of dependent tokens leads to inconsistent predictions, as demonstrated by the distorted local patterns in Fig.~\ref{fig:teaser} (b).
For visual data, such dependencies are naturally correlated with spatial distances\textemdash while locally adjacent tokens exhibit strong dependencies, spatially distant tokens often have weak correlations. This motivates us to reconsider how to organize tokens for generation: by identifying spatially distant tokens with weak correlations, we can group them for simultaneous prediction. Such non-local grouping allows us to maintain sequential generation for strongly dependent local tokens while enabling parallel generation across different spatial regions. Moreover, we observe that initial tokens in each local region play a crucial role in establishing the global structure - generating them in parallel could lead to conflicting structures across regions, such as repeated parts in different regions without global coordination(see middle row of Fig.~\ref{fig:exp_vis}). Therefore, the initial tokens in each local region should be generated sequentially to establish the global visual structure.

Based on these insights, we propose a simple yet effective approach for parallel generation in autoregressive visual models. Our key idea is to identify and group weakly dependent visual tokens for simultaneous prediction while maintaining sequential generation for strongly dependent ones. 
To achieve this, we first divide the image into local regions and generate their initial tokens sequentially to establish global context, then perform parallel generation by identifying and grouping tokens at corresponding positions across spatially distant regions. The process is illustrated in Fig.~\ref{fig:teaser} (a).
Our approach can be seamlessly implemented within standard autoregressive transformers through a reordering mechanism, with a few learnable token embeddings to facilitate the transition between sequential and parallel generation modes. By ensuring each prediction step has access to all previously generated tokens across regions, we maintain the autoregressive property and preserve global context modeling capabilities. 
With non-local parallel generation, our approach significantly reduces the number of inference steps and thereby accelerates generation, while maintaining comparable visual quality through careful token dependency handling.
% Through the non-local parallel generation, our approach significantly reduces the number of inference steps and thereby accelerates generation, while maintaining comparable visual quality due to the careful handling of token dependencies.

We verify the effectiveness of our approach on both image and video generation tasks using ImageNet~\cite{imagenet} and UCF-101~\cite{soomro2012ucf101} datasets. For image generation, our method achieves around 3.9$\times$ fewer generation steps and 3.6$\times$ actual inference-time speedup with comparable generation quality. With more aggressive parallelization, we achieve around 11.3$\times$ reduction in steps and 9.5$\times$ actual speedup with minimal quality drop (within 0.7 FID for image and 10 FVD for video). The qualitative comparison of generation results between our method and the baseline is shown in Fig.~\ref{fig:intro_vis}. 
The experiments demonstrate the effectiveness of our approach across different visual domains and its compatibility with various tokenizers like VQGAN~\cite{esser2020taming} and MAGVIT-v2~\cite{yu2023language}.

In summary, we propose a simple yet effective parallelized autoregressive visual generation approach that carefully handles token dependencies. Our key idea is to identify and group weakly dependent tokens for simultaneous prediction while maintaining sequential generation for strongly dependent ones. Our approach can be seamlessly integrated into standard autoregressive models without architectural modifications. Through extensive experiments with different visual domains and tokenization methods, we demonstrate considerable speedup while preserving generation quality, making autoregressive visual generation more practically usable for real-world applications.

\section{Related Work}
\label{sec:related_work}

% \vspace{1pt}
% \noindent\textbf{Autoregressive Visual Generation.} 
% Autoregressive modeling has been explored in visual generation for years. Early approaches~\cite{van2016conditional,van2016pixel,salimans2017pixelcnn++} model images as sequences of pixels, predicting each pixel conditioned on all previous ones. Current approaches typically follow a two-stage paradigm: in the first stage, visual data is compressed into compact sequences of tokens through discrete tokenizers~\cite{vqvae,esser2020taming,yu2023language}, where each token represents a learned visual pattern from a codebook; in the second stage, a transformer~\cite{vaswani2017attention} is trained to model the joint distribution of these tokens by predicting them autoregressively following a raster scan order~\cite{ramesh2021dalle,yu2022scaling,rq}. This token-based paradigm has been successfully extended to video generation~\cite{kondratyuk2023videopoet,yan2021videogpt,wang2024loong}, where tokens from different frames are predicted sequentially. However, the strictly sequential generation process leads to slow inference speed that scales with sequence length.
\noindent\textbf{Autoregressive Visual Generation.}
Autoregressive modeling has been explored in visual generation for years, from early pixel-based approaches~\cite{van2016conditional,van2016pixel,salimans2017pixelcnn++} to current token-based methods. Modern approaches typically follow a two-stage paradigm: first compressing visual data into compact token sequences through discrete tokenizers~\cite{vqvae,esser2020taming,yu2023language}, then training a transformer~\cite{vaswani2017attention} to predict these tokens autoregressively in raster scan order~\cite{ramesh2021dalle,yu2022scaling,rq}. This paradigm has been successfully extended to video generation~\cite{kondratyuk2023videopoet,yan2021videogpt,wang2024loong}, where tokens from different frames are predicted sequentially. However, the strictly sequential generation process leads to slow inference speed that scales with sequence length.

\vspace{2pt}
\noindent\textbf{Parallel Prediction in Sequential Generation.}
Various approaches have been proposed to accelerate sequential generation. In language modeling, speculative decoding~\cite{leviathan2023fast,chen2023accelerating,li2024eagle} employs a draft model to generate candidate tokens for main model verification, while Jacobi decoding~\cite{song2021accelerating,kou2024cllms} enables parallel generation through iterative refinement. In visual generation, MaskGIT~\cite{chang2022maskgit} adopts a non-autoregressive approach with BERT-like masked modeling strategies, taking a different modeling paradigm from traditional autoregressive generation. VAR~\cite{tian2024visual} proposes next-scale prediction that progressively generates tokens at increasing resolutions, though requiring specialized multi-level tokenizers and longer token sequences. In contrast, our approach enables efficient parallel generation while preserving the autoregressive property and model simplicity, readily applicable to various visual tasks without specialized architectures or additional models.

\section{Method}
\label{sec:method}

\begin{figure*}[htbp]
   \centering
   \includegraphics[width=\linewidth]{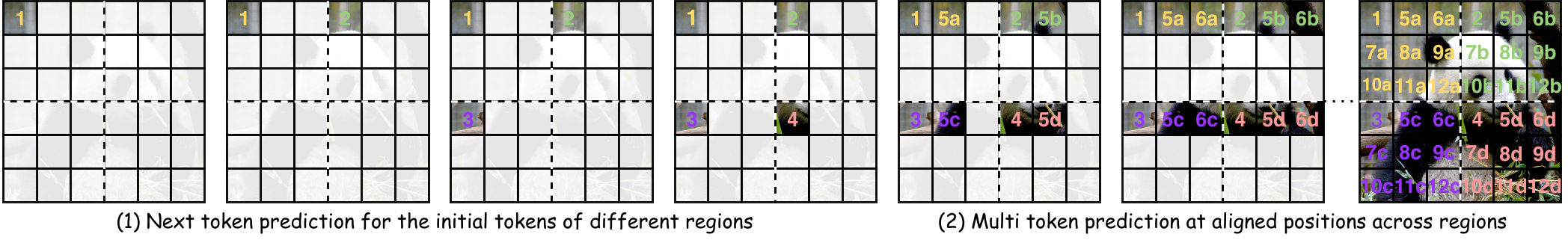}
    \vspace{-20pt}
   \caption{\textbf{Illustration of our non-local parallel generation process.} Stage 1: sequential generation of initial tokens (1-4) for each region (separated by dotted lines) to establish global structure. Stage 2: parallel generation at aligned positions across different regions (e.g., 5a-5d), then moving to next aligned positions (6a-6d, 7a-7d, etc.) for parallel generation. Same numbers indicate tokens generated in the same step, and letter suffix (a,b,c,d) denotes different regions .}
   \label{fig:method}
   \vspace{-13pt}
\end{figure*}

In this section, we present our approach for parallelized visual autoregressive generation. We first discuss the relationship between token dependencies and parallel generation in Sec.~\ref{sec:3.1}. Based on these insights, we propose our parallel generation approach in Sec.~\ref{sec:3.2}. Finally, we present the model architecture and implementation details that realize this process within autoregressive transformers in Sec.~\ref{sec:3.3}.

\subsection{Token Dependencies and Parallel Generation}
\label{sec:3.1}

Standard autoregressive models adopt token-by-token sequential generation, which significantly limits generation efficiency. To improve efficiency, we explore the possibility of generating multiple tokens in parallel. However, a critical question arises: \textit{which tokens can be generated in parallel without compromising generation quality?} In this section, we analyze the relationship between token dependencies and parallel generation through pilot studies, providing guidance for designing parallelized autoregressive visual generation models.

\vspace{2pt}
\noindent\textbf{Pilot Study.} In language modeling, researchers have attempted to group adjacent tokens for multi-token prediction~\cite{wang2018semi,stern2018blockwise,leviathan2023fast,chen2023accelerating,li2024eagle,song2021accelerating,kou2024cllms}. However, our pilot study reveals that directly predicting adjacent tokens leads to significant quality degradation in visual generation (see Fig.~\ref{fig:teaser}(b) and Tab.~\ref{tab:token_design} (d)). In autoregressive generation, each token is generated through sampling strategies (e.g., top-k) to maintain diversity. When generating multiple tokens in parallel, these tokens need to be sampled independently. However, for adjacent visual tokens with strong dependencies, their joint distribution cannot be factorized into independent distributions, as each token is heavily influenced by its neighbors. The impact of such independent sampling is clearly demonstrated in the figure, where generating adjacent tokens in parallel leads to inconsistent local structures like distorted tiger faces and fragmented zebra stripes, as tokens are sampled without considering their neighbors' decisions.

\vspace{2pt}
\noindent\textbf{Design Principles.} These observations suggest that parallel generation should focus on weakly correlated tokens to minimize the impact of independent sampling. For visual tokens, dependencies naturally decrease with spatial distance - tokens from distant regions typically have weaker correlations than adjacent ones. This motivates us to perform parallel generation across distant regions rather than within local neighborhoods.
However, we find that not all distant tokens can be generated in parallel. The initial tokens of each regions are particularly crucial as they jointly determine the global image structure. Parallel generation of these initial tokens, despite their spatial distances, could lead to conflicting global decisions, resulting in issues like repeated patterns or incoherent patches across regions (see the middle row in Fig.~\ref{fig:exp_vis}).

Based on these insights, we propose three key design principles for parallelized autoregressive generation: 1) generate initial tokens for each region sequentially to establish proper global structure; 2) maintain sequential generation within local regions where dependencies are strong; and 3) enable parallel generation across regions where dependencies are weak through proper token organization.

\begin{figure*}[htbp]
   \centering
   \includegraphics[width=\linewidth]{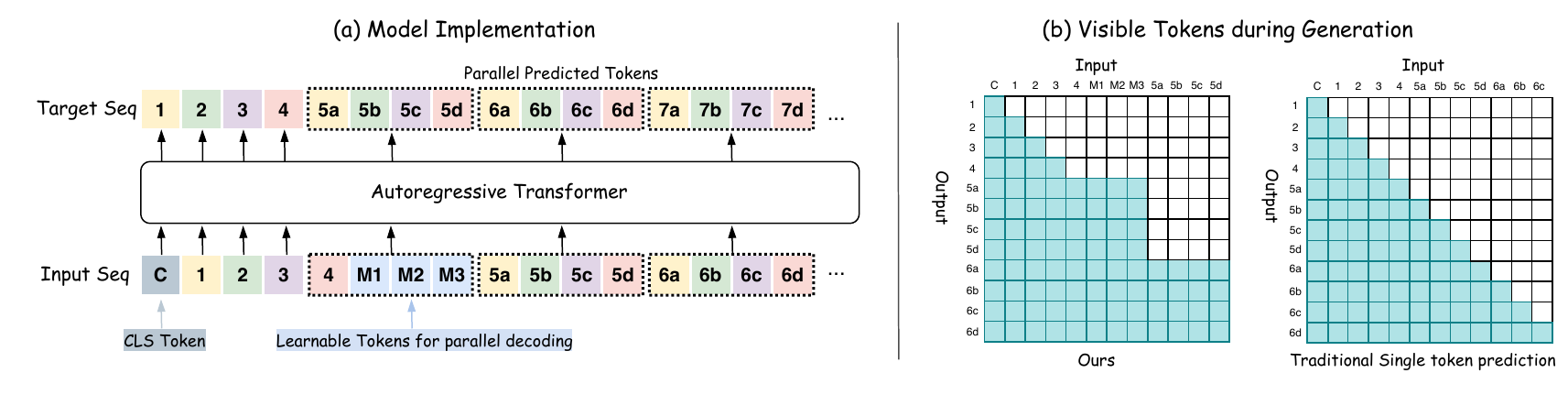}
   \vspace{-30pt}
       \caption{\textbf{Overview of our parallel autoregressive generation framework.} \textbf{(a) Model implementation.} The model first generates initial tokens sequentially [1,2,3,4], then uses learnable tokens [M1,M2,M3] to help transition into parallel prediction mode.
       % For predicting each group (e.g., [6a-6d]), the model takes all previous tokens as input. 
       % (b) Comparison of visible context between our parallel prediction approach (left) and traditional single-token prediction (right). The colored cells indicate available context during generation, highlighting how parallel prediction reduces available context, which is partially compensated by allowing full attention within each prediction group.
       \textbf{(b) Comparison of visible context between our parallel prediction approach }(left) \textbf{and traditional single-token prediction} (right). The colored cells indicate available context during generation. In traditional AR, when predicting token $6d$, the model can access all previous tokens including $6a-6c$. Without full attention, our parallel approach would limit each token (e.g., $6b$) to only see tokens up to the same position in the previous group (e.g., up to $5b$). We enable group-wise full attention to allow access to the entire previous group.
       }
   \vspace{-10pt}
   \label{fig:implementation}
\end{figure*}

\subsection{Non-Local Parallel Generation} \label{sec:3.2}

Based on the above principles, we propose our approach that enables parallel token prediction while maintaining autoregressive properties.
The process is illustrated in Fig.~\ref{fig:method}.

\vspace{2pt}
\noindent\textbf{Cross-region Token Grouping.} 
Let $\{v_i\}_{i=1}^{H \times W}$ denote a sequence of visual tokens arranged in a $H \times W$ grid. We first partition the token grid into $M \times M$ regions. Each region contains $k := \Big(\frac{H}{M} \times \frac{W}{M}\Big)$ tokens. We then group tokens at corresponding positions across different regions. Let $v^{(r)}_{j}$ denote the token at position $j$ in region $r$, where $r \in \{1,...,M^2\}$ and $j \in \{1,...,k\}$. 
We then organize these tokens into groups based on their corresponding positions across regions:
\begin{equation}
\small
   \Big\{[v^{(1)}_{1},\cdots,v^{(M^2)}_{1}], [v^{(1)}_{2},\cdots,v^{(M^2)}_{2}], \cdots, [v^{(1)}_{k},\cdots.,v^{(M^2)}_{k}]\Big\}.
\end{equation}
This organization groups together tokens at the same relative position across different regions, facilitating our parallel generation process.

\vspace{2pt}
\noindent\textbf{Stage 1: Sequential Generation of Initial Tokens of Each Region.} 
We first generate one initial token for each region sequentially (marked as ``1-4'' in Fig.~\ref{fig:method}) to establish the global context. As shown in Fig.~\ref{fig:method} (1), we start with the top-left region and generate the initial token for each region by sampling from the conditional probability distribution:
\begin{equation}
v^{(i)}_1 \sim \mathbb{P}(v^{(i)}_1 | v^{(<i)}_1), \quad i \in \{1,...,M^2\},
\end{equation}
where $v^{(i)}_1$ denotes the initial token of the $i$-th region. Since the number of regions ($M^2$) is small and fixed, this sequential generation introduces minimal overhead while providing crucial global context for subsequent parallel generation.

\vspace{2pt}
\noindent\textbf{Stage 2: Parallel Generation of Cross-region Tokens.}
After initializing tokens for all regions, we proceed with parallel generation of the remaining tokens. As illustrated in Fig.\ref{fig:method} (2), at each step, we identify the next position $j$ within each region following a raster scan order and simultaneously predict tokens at this position across all regions (e.g., tokens 5a-5d are generated in parallel). The parallel generation at each step can be formulated as:
\begin{equation}
\{v^{(r)}_{j}\}_{r=1}^{M^2} \sim \mathbb{P}(\{v^{(r)}_{j}\}_{r=1}^{M^2} | v_{<j}),
\end{equation}
where $\{v^{(r)}_{j}\}_{r=1}^{M^2}$ represents the set of tokens at position $j$ across all regions to be generated in parallel, and $v_{<j}$ includes both initial tokens and tokens from previous parallel steps. For example, with $M=2$ on a $24 \times 24$ token grid, after generating 4 initial tokens sequentially, we predict $M^2=4$ tokens in parallel at each subsequent step, reducing the total number of generation steps from 576 to 147 (i.e., $4 + \frac{576-4}{4}$). While enabling parallel prediction, our approach maintains the autoregressive property as each prediction is still conditioned on all previous tokens. The key difference is that tokens at corresponding positions across regions, which exhibit weak dependencies, are now generated simultaneously instead of sequentially.

\subsection{Model Architecture  Details} \label{sec:3.3}

We illustrate our parallel generation framework using a standard autoregressive transformer for class-conditioned image generation.

\vspace{2pt}
\noindent\textbf{Framework Implementation.} As shown in Fig.~\ref{fig:implementation} (a), our model architecture consists of an autoregressive transformer that processes the input sequence and generates visual tokens. The input sequence begins with a class token $(C)$ followed by visual tokens to be generated. To achieve $n$-token parallel prediction, we design a special sequence structure with three distinct parts: 1) initial sequential tokens [1,2,...,$n$] that are generated one at a time, 2) a transition part with $n-1$ learnable tokens [M1,M2,M3] that helps the model enter parallel prediction mode, and 3) subsequent token groups that are predicted $n$ tokens at a time (e.g., [$5a,5b,5c,5d$], [$6a,6b,6c,6d$]). For predicting each group, the model takes all previous tokens as input while maintaining a fixed offset of $n$ tokens between input and target sequences. The learnable tokens share the same dimension as regular tokens for seamless integration. To maintain spatial relationships under our reordered sequence, we employ 2D Rotary Position Embedding (RoPE)~\citep{su2024roformer}, which preserves each token's original spatial position information regardless of its sequence position. The above designs enable parallel prediction while preserving the standard autoregressive transformer architecture.

\vspace{2pt}
\noindent\textbf{Group-wise Bi-directional Attention with Global Autoregression.} 
Our framework combines sequential generation of initial tokens with parallel generation of subsequent token groups. As illustrated in Fig.\ref{fig:implementation} (b), in traditional autoregressive models, when predicting token $6d$, the model can access all previous tokens including $6a-6c$. However, naive parallel generation with causal masking would restrict each token (e.g., $6b$) to only see tokens up to the same position in the previous group (e.g., up to $5b$), limiting the available context. To address this limitation while maintaining parallelism, we enable bi-directional attention within each prediction group while preserving causal attention between groups. This allows each token in the current group to access the entire previous group as context (e.g., all tokens [$5a-5d$] are visible when predicting any token in [$6a-6d$]). This design enriches the local context for parallel prediction while maintaining the global autoregressive property, ensuring compatibility with standard optimizations like KV-cache.

\vspace{2pt}
\noindent\textbf{Extension to Video Generation.} Our parallel generation framework can be naturally extended to video generation. The tokenization process reduces both spatial and temporal dimensions, resulting in tokens arranged in a $T \times H \times W$ grid, where each latent frame aggregates information from multiple input frames. We treat these temporally compressed tokens similarly to image tokens and apply our parallel generation strategy along the spatial dimensions, with the only modification being the use of 3D position embeddings. 
While we also explored parallel generation along the temporal dimension, we found it less effective than spatial parallelization. This is because temporal dependencies exhibit stronger sequential characteristics that are fundamental to video coherence, making them less suitable for parallel prediction compared to spatial relationships. The exploration of effective temporal parallel strategies remains as future work.
% While we also explored parallel generation along the temporal dimension, we found it less effective than spatial parallelization, as temporal dependencies follow a natural sequential order that is less amenable to reordering compared to spatial relationships. Further exploration of temporal parallel strategies remains as future work.

\section{Experiments}
\label{sec:exp}

\subsection{Experimental Setup}

\begin{table}[t]
\setlength{\tabcolsep}{4pt}
\centering
\begin{tabular*}{\linewidth}{@{\extracolsep{\fill}}lcccc@{}}
\toprule
\textbf{Model} & \textbf{Params} & \textbf{Layers} & \textbf{Hidden} & \textbf{Heads} \\
\midrule
\our-L & 343M & 24 & 1024 & 16 \\
\our-XL & 775M & 36 & 1280 & 20 \\
\our-XXL & 1.4B & 48 & 1536 & 24 \\
\our-3B & 3.1B & 24 & 3200 & 32 \\
\bottomrule
\end{tabular*}
\vspace{-3mm}
\caption{\textbf{Model sizes and architecture configurations of \our.} The configurations are following previous works~\citep{radford2019language,touvron2023llama,open_llama_3b,sun2024autoregressive}.}
\label{tab:model_scaling}
\vspace{-7mm}
\end{table}

\begin{figure*}[t]
   \centering
   \includegraphics[width=.9\linewidth]{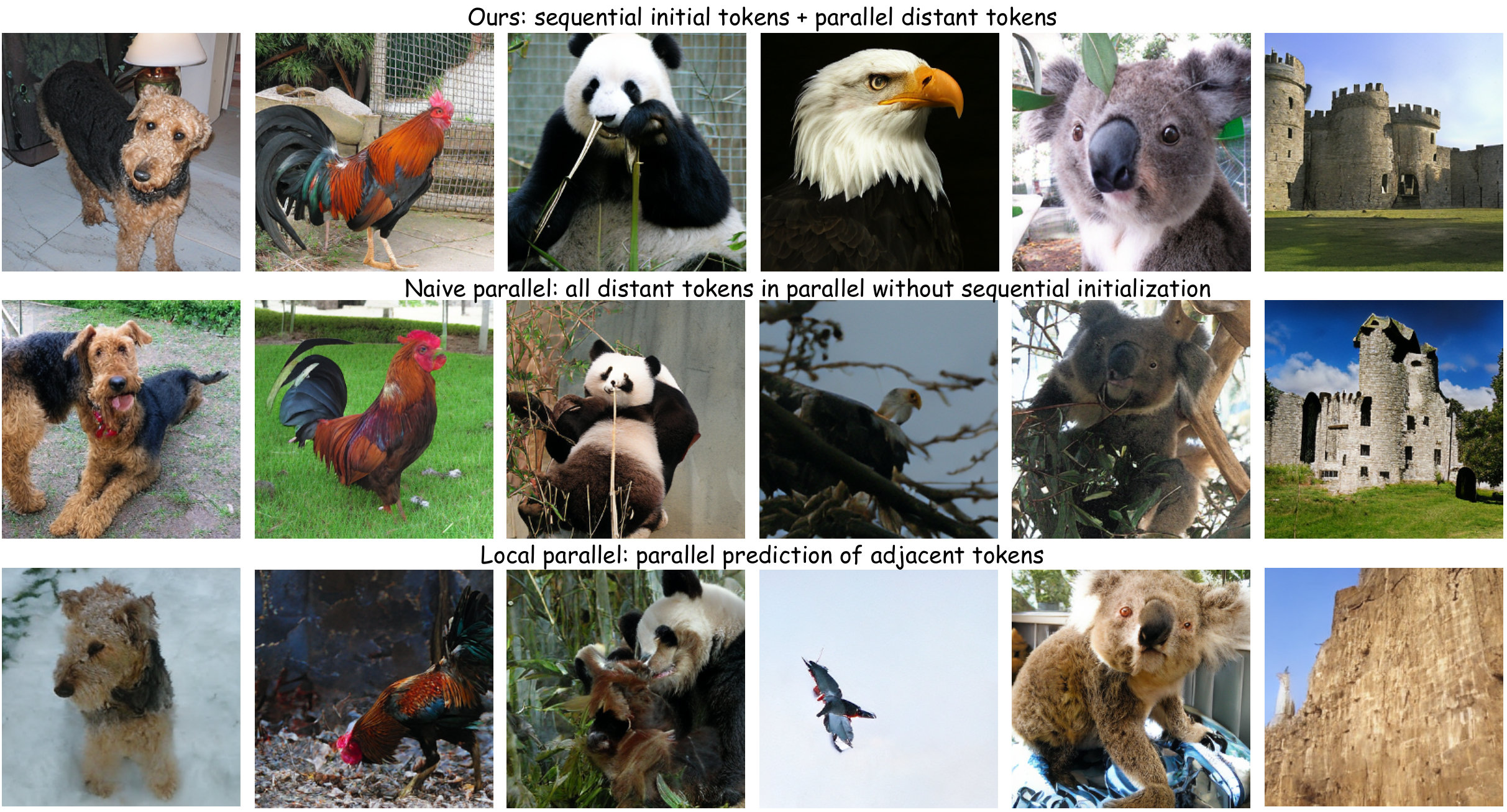}
   \vspace{-8pt}
   \caption{\textbf{Qualitative comparison of parallel generation strategies.} \textbf{Top:} Our method with sequential initial tokens followed by parallel distant token prediction produces high-quality and coherent images. \textbf{Middle:} Direct parallel prediction without sequential initial tokens leads to inconsistent global structures. \textbf{Bottom:} Parallel prediction of adjacent tokens results in distorted local patterns and broken details.}
   \vspace{-8pt}
   \label{fig:exp_vis}
\end{figure*}

\begin{table*}[t]
\centering
\scalebox{0.84}{
\begin{tabular}{c|lc|cccccc}
\toprule
Type & Model & \#Para. & FID$\downarrow$ & IS$\uparrow$ & Precision$\uparrow$ & Recall$\uparrow$ & Steps & Time(s)$\downarrow$ \\
\midrule
\multirow{3}{*}{GAN}   & BigGAN~\cite{biggan}  & 112M   & 6.95  & 224.5       & 0.89 & 0.38 &1&$-$\\
 & GigaGAN~\cite{gigagan}  & 569M    & 3.45  & 225.5       & 0.84 & 0.61  & 1 &$-$\\
 & StyleGan-XL~\cite{stylegan-xl} & 166M    & 2.30  & 265.1       & 0.78 & 0.53   & 1 &0.08\\
\midrule
\multirow{4}{*}{Diffusion} & ADM~\cite{dhariwal2021diffusion}  & 554M       & 10.94 & 101.0        & 0.69 & 0.63    & 250 &44.68\\
 & CDM~\cite{cdm}   & $-$       & 4.88  & 158.7       & $-$  & $-$   & 8100 &$-$\\
 & LDM-4~\cite{rombach2022high} & 400M     & 3.60  & 247.7       & $-$  & $-$  & 250 &$-$\\
 & DiT-XL/2~\cite{peebles2023scalable}  & 675M  & 2.27  & 278.2       & 0.83 & 0.57   & 250 &11.97\\
\midrule
\multirow{1}{*}{Mask} & MaskGIT~\cite{chang2022maskgit}  & 227M   & 6.18  & 182.1        & 0.80 & 0.51  & 8 &0.13\\
\midrule
\multirow{1}{*}{VAR} & VAR-d30~\cite{tian2024visual}  & 2B   & 1.97  & 334.7        & 0.81 & 0.61  & 10 &0.27\\
\midrule
\multirow{1}{*}{MAR} & MAR~\cite{li2024autoregressive}  & 943M   & 1.55  & 303.7        & 0.81 & 0.62  & 64 &28.24\\
\midrule
\multirow{7}{*}{AR} & VQGAN~\cite{esser2020taming} & 227M & 18.65 & 80.4         & 0.78 & 0.26    & 256 &5.05\\
 & VQGAN~\cite{esser2020taming}    & 1.4B   & 15.78 & 74.3   & $-$  & $-$    & 256 &5.05 \\
 & VQGAN-re~\cite{esser2020taming}  & 1.4B  & 5.20  & 280.3  & $-$  & $-$     & 256 &6.38\\
 & ViT-VQGAN~\cite{vit-vqgan} & 1.7B & 4.17  & 175.1  & $-$  & $-$       & 1024 & $>$6.38\\
 & ViT-VQGAN-re~\cite{vit-vqgan}& 1.7B  & 3.04  & 227.4  & $-$  & $-$     & 1024 &$>$6.38\\
 & RQTran.~\cite{rq}       & 3.8B  & 7.55  & 134.0  & $-$  & $-$     & 256 &5.58\\
 & RQTran.-re~\cite{rq}    & 3.8B & 3.80  & 323.7  & $-$  & $-$    & 256 &5.58\\
\midrule
\multirow{4}{*}{AR} & LlamaGen-L~\cite{sun2024autoregressive}   & 343M & 3.07 & 256.1 & 0.83 & 0.52 &576&12.58\\
 & LlamaGen-XL~\cite{sun2024autoregressive}   & 775M & 2.62 & 244.1 & 0.80 & 0.57  &576&18.66\\
 & LlamaGen-XXL~\cite{sun2024autoregressive}  & 1.4B & 2.34 & 253.9 & 0.80 & 0.59  &576&24.91\\
 & LlamaGen-3B~\cite{sun2024autoregressive}  & 3.1B & 2.18 & 263.3 & 0.81 & 0.58  &576&12.41\\
\midrule
\multirow{6}{*}{AR} 
 & \our-L-4$\times$   & 343M &  3.76& 218.9  & 0.84 &  0.50 & 147 &3.38\\
 & \our-XL-4$\times$   & 775M & 2.61 &  259.2&   0.82 & 0.56 & 147 &4.94\\

 & \our-XXL-4$\times$  & 1.4B & 2.35 & 263.2 & 0.82 & 0.57& 147 &6.84\\ 
 & \our-3B-4$\times$  & 3.1B &  2.29&  255.5& 0.82 & 0.58 & 147 &3.46\\ 
 \cdashlinelr{2-9} % \cdashline{2-9}
 & \our-XXL-16$\times$ & 1.4B & 3.02 & 270.6 & 0.81 & 0.56 &  51&2.28\\
 & \our-3B-16$\times$ & 3.1B &  2.88&  262.5&  0.82& 0.56 &  51&1.31\\

\bottomrule
\end{tabular}
}
\vspace{-5pt}
\caption{\textbf{Class-conditional image generation on ImageNet 256$\times$256 benchmark}. 
% Models are grouped by types: GAN (one-shot generation), Diffusion (iterative refinement), Mask (mask token prediction), VAR (multi-scale generation), MAR (continuous mask token prediction), and AR (autoregressive generation).
``$\downarrow$'' or ``$\uparrow$'' indicate lower or higher values are better.
``-re'' means using rejection sampling. %``cfg'' means using classifier-free guidance.
\our-4$\times$ and \our-16$\times$ means generating 4 and 16 tokens per step in parallel, respectively. 
}
\label{tab:main}
\vspace{-10pt}
% \vspace{-5mm}
\end{table*}

\noindent\textbf{Image Generation.} 
For fair comparison with existing token-by-token autoregressive visual generation methods, we adopt similar settings as~\cite{sun2024autoregressive}, using a VQGAN tokenizer~\cite{esser2020taming} with a 16,384 codebook size and 16$\times$ downsampling ratio. Models are trained on ImageNet-1K~\cite{imagenet} for 300 epochs, with 384$\times$384 images tokenized into 24$\times$24 sequences. We evaluate on the ImageNet validation set at 256$\times$256 resolution using FID~\cite{fid} as the primary metric, complemented by IS and Precision/Recall~\cite{precision_recall}. We experiment with model sizes from 343M to 3.1B parameters (Tab.\ref{tab:model_scaling}), reporting both generation steps and latency time.

\vspace{1pt}
\noindent\textbf{Video Generation.}
We evaluate on the UCF-101~\citep{soomro2012ucf101} dataset using MAGVIT-v2 tokenizer~\cite{yu2023language} reproduced by ~\cite{ren2025nbp}. Each 17-frame video (128$\times$128 resolution) is compressed by 8$\times$ spatially and 4$\times$ temporally into a $5\times16\times16$ token sequence (1280 tokens per video). For fair comparison, we implement both next-token prediction and our parallel generation approach using the same architecture. The position of video codes is encoded via 3D positional embeddings. Our reproduced MAGVIT-v2 tokenizer uses a 64K visual vocabulary instead of the original 262K to facilitate model training. We use Fréchet Video Distance (FVD)~\cite{unterthiner2018towards} to evaluate generation quality. 

Detailed training configurations for both video and image generation are provided in the supplementary material.

\subsection{Main Results}

\subsubsection{Image Generation}
% We evaluate our approach on class-conditional image generation using the ImageNet~\cite{imagenet} 256$\times$256 benchmark. 
Tab.~\ref{tab:main} presents comprehensive comparisons of class-conditional image generation with various state-of-the-art methods, including 
GAN~\cite{biggan,gigagan,stylegan-xl}(one-shot generation), Diffusion~\cite{dhariwal2021diffusion,cdm,rombach2022high,peebles2023scalable} (iterative denoising), Mask~\cite{chang2022maskgit} (mask token prediction), VAR~\cite{tian2024visual} (next-scale prediction), MAR~\cite{li2024autoregressive} (continuous mask token prediction), and AR~\cite{sun2024autoregressive,vit-vqgan,esser2020taming} (autoregressive generation). Our \our achieves competitive performance while maintaining faster inference speed than most state-of-the-art models.
Specifically, when comparing with representative models from different categories, our method shows advantages. Compared with the mask-based method MaskGIT~\cite{chang2022maskgit}, our method achieves substantially better generation quality (FID 2.29 vs. 6.18) despite requiring more steps. For VAR~\cite{tian2024visual}, while it achieves slightly better FID (1.97 vs. 2.29), our method maintains a simpler framework with fewer tokens per image and preserves the pure autoregressive nature, making it more flexible for multi-modal integration.
% Compared with MaskGIT~\cite{chang2022maskgit}, although it requires fewer generation steps, our method achieves substantially better generation quality (FID 2.29 vs. 6.18). Compared with VAR~\cite{tian2024visual}, while our FID is slightly higher (2.29 vs. 1.97), our method maintains a simpler framework that requires fewer tokens per image and preserves the pure autoregressive nature, making it more flexible and easier to integrate with other modalities.
 
Compared to our baseline model LlamaGen~\cite{sun2024autoregressive}, \our achieves 3.9$\times$ reduction in generation steps (147 vs.\ 576) and 3.58$\times$ speedup in wall-clock time (3.46s vs.\ 12.41s) while maintaining comparable quality (FID 2.29 vs.\ 2.18). With more aggressive parallelization, \our-3B-16$\times$ further accelerates generation to 1.31s (9.5$\times$ speedup) with only 0.7 FID degradation compared to the baseline, demonstrating the effectiveness of our parallel generation strategy in balancing efficiency and quality.
% \textcolor{red}{[JS: better to first say our method is faster than most of the SOTA methods. Then specifically discuss MaskGIT and VAR sperately. Highlight our method generates higher-quality image than MaskGIT. And our method performs comparably with VAR but maintains simplicity and autoregressive nature, easy to integrate with other modalities. ]}

% Our largest model \our-3B-4 achieves competitive performance (FID 2.29) compared to recent state-of-the-art methods like DiT-XL/2 (FID 2.27) and VAR-d30 (FID 1.97). More importantly, compared to our direct baseline LlamaGen, our parallel prediction approach achieves 3.9$\times$ reduction in generation steps (147 vs.\ 576) and 3.58$\times$ speedup in wall-clock time (3.46s vs.\ 12.41s) while maintaining comparable quality (FID 2.29 vs.\ 2.18, precision 0.82 vs.\ 0.81, recall 0.58 vs.\ 0.58). With more aggressive parallelization, \our-3B-16 further reduces the steps to 51, demonstrating the effectiveness of our parallel generation strategy in balancing generation efficiency and quality.

\begin{table}[t]
\centering
\begin{adjustbox}{max width=\linewidth}
\begin{tabular}{l|lc|ccc}
\toprule
Type & Method & \#Param & FVD$\downarrow$ & Steps & Time(s) \\
\midrule
% \multirow{1}{*}{GAN} & DVD-GAN~\citep{clark2019adversarial} & N/A & - & - & - \\ \midrule
\multirow{3}{*}{Diffusion} & VideoFusion~\citep{luo2023videofusion} & N/A & 173 & - & -\\
 & Make-A-Video~\cite{make-a-video} & N/A & 81.3 & - & -\\
 & HPDM-L~\citep{skorokhodov2024hierarchical}  & 725M & 66.3 & - & -\\
\midrule
\multirow{2}{*}{Mask.} 
 % & Phenaki~\cite{villegas2022phenaki} & 227M & - & -& - \\
 & MAGVIT~\cite{yu2023magvit} & 306M & 76 & - & -\\
 & MAGVIT-v2~\cite{yu2023language} & 840M & 58 & - & -\\
\midrule
\multirow{4}{*}{AR} 
 % & LVT~\citep{rakhimov2020latent} & 50M & - & - & -\\
 % & ViTrans~\citep{weissenborn2019scaling} & 373M & - & - & -\\
 & CogVideo~\cite{hong2022cogvideo} & 9.4B & 626 & - & -\\
 % & ViVQVAE~\citep{walker2021predicting} & N/A & - & - & -\\
 & TATS~\citep{ge2022long} & 321M & 332 & - & -\\
 & OmniTokenizer~\citep{wang2024omnitokenizer} & 650M & 191 & 5120 & 336.70\\
 & MAGVIT-v2-AR~\cite{yu2023language} & 840M & 109 & 1280 & -\\
\midrule
\multirow{3}{*}{AR} &\our-1$\times$ & 792M & 94.1 & 1280 & 43.30\\
 &\our-4$\times$ & 792M & 99.5 & 323 & 11.27\\
 &\our-16$\times$ & 792M & 103.4 & 95 & 3.44\\
\bottomrule
\end{tabular}
\end{adjustbox}
\vspace{-8pt}
\caption{\textbf{Comparison of class-conditional video generation methods on UCF-101 benchmark.} 
FVD measures generation quality, where lower values ($\downarrow$) indicate better performance. \our-1$\times$ represents our token-by-token baseline, while \our-4$\times$ and \our-16$\times$ indicate our parallel generation variants with different speedup ratios, achieving competitive FVD scores with significantly reduced generation steps and wall-clock time.}
\vspace{-18pt}
\label{tab:main_results}
\end{table}

\subsubsection{Video Generation}
We evaluate our approach on the UCF-101~\cite{soomro2012ucf101} dataset for class-conditional video generation. Table~\ref{tab:main_results} shows comparisons with various state-of-the-art methods across different categories. Among recent works, MAGVIT-v2~\cite{yu2023language} achieves strong performance with an FVD of 58 using masked token prediction, while its autoregressive variant MAGVIT-v2-AR obtains an FVD of 109 with 1280 generation steps. Our next-token-prediction baseline (\our-1$\times$) achieves a competitive FVD of 94.1,  demonstrating the effectiveness of our implementation. More importantly, our parallel generation variants significantly reduce both generation steps and wall-clock time while maintaining comparable quality. Specifically, \our-4$\times$ reduces the generation steps from 1280 to 323 with minimal FVD increase (99.5 vs. 94.1), achieving 3.8$\times$ speedup (11.27s vs. 43.30s). Further parallelization with \our-16$\times$ achieves 12.6$\times$ speedup (3.44s vs. 43.30s) with 103.4 FVD, while reducing generation steps to 95. Due to space limit, we provide visualization results of video generation in supplementary materials.

\subsection{Ablation Study}
In this section, we conduct comprehensive ablation studies to investigate the effectiveness of our key design choices on the ImageNet 256$\times$256 validation set (Tab.~\ref{tab:token_design}). Unless specified, we use the \our-XL model with parallel group size n=4 as default setting.
% \textcolor{red}{[JS: mention the purpose of ablation study, like investigating the importance of each component or different design choices. or for better understanding xxx. ]}

\begin{table}[t]
\centering
\begin{subtable}[t]{\linewidth}
\centering
\begin{tabular}{c|ccc}
\toprule
 & FID$\downarrow$ & IS$\uparrow$ & steps$\downarrow$ \\
\midrule
 w/o & 3.67 & 221.36 &  144\\
w & \textbf{2.61} & 259.17  & 147 \\
\bottomrule
\end{tabular}
\caption{\textbf{Importance of initial sequential token generation.} Sequential generation of initial tokens improves FID by 1.06 with negligible step increase.}
\end{subtable}

% \vspace{4mm}

\begin{subtable}[t]{\linewidth}
\centering
\begin{tabular}{c|ccc}
\toprule
n & FID$\downarrow$ & IS$\uparrow$ & steps$\downarrow$ \\
\midrule
1 & \textbf{2.34} & 253.90 & 576 \\
4 & 2.35 & 263.24 & 147 \\
16 & 3.02 & 270.57 & 51 \\
\bottomrule
\end{tabular}
\caption{\textbf{Number of parallel predicted tokens} (\our-XXL). n=1 is the token-by-token baseline. n=4 reduces steps by 4$\times$ with similar FID (2.35 vs. 2.34), while n=16 reduces steps by 11.3$\times$ at the cost of 0.67 FID.}
\end{subtable}

% \vspace{4mm}

\begin{subtable}[t]{\linewidth}
\centering
\begin{tabular}{c|ccc}
\toprule
attn & FID$\downarrow$ & IS$\uparrow$ & steps$\downarrow$ \\
\midrule
causal & 3.64 & 228.08 & 147 \\
full & \textbf{2.61} & 259.17 & 147 \\
\bottomrule
\end{tabular}
\caption{\textbf{Attention pattern between parallel tokens.} Full attention allows complete context access from previous parallel groups (vs. causal attention's limited access), bringing 1.03 FID improvement.}% while maintaining the same generation steps.}
\end{subtable}

% \vspace{4mm}

\begin{subtable}[t]{\linewidth}
\centering
\begin{tabular}{lc|ccc}
\toprule
order&pattern & FID$\downarrow$ & IS$\uparrow$ & steps$\downarrow$ \\
\midrule
raster &one  & 2.62 & 244.08 & 576 \\
distant &one & 2.64 & 262.72  & 576 \\
raster &multi  & 5.64 & 265.46 & 147 \\
distant &multi& \textbf{2.61} & 259.17 & 147 \\
\bottomrule
\end{tabular}
\caption{\textbf{Comparison of different scan orders under single-token and multi-token prediction.} Our region-based distant ordering shows similar performance with raster scan in single-token setting, but significantly outperforms in multi-token prediction (2.61 vs. 5.64 FID).}
\end{subtable}

% \vspace{4mm}

\begin{subtable}[t]{\linewidth}
\centering
\begin{tabular}{c|ccc}
\toprule
Params & FID$\downarrow$ & IS$\uparrow$ & steps \\
\midrule
343M &  3.76& 218.92 &147 \\
775M & 2.61 & 259.17 &147  \\
1.4B & 2.35 & 263.24 &147 \\
3.1B & \textbf{2.29}&  255.46 &147 \\
\bottomrule
\end{tabular}
\caption{\textbf{Scaling of model size} (4$\times$ parallel). Generation quality steadily improves with more parameters, from 343M (FID 3.76) to 3.1B (FID 2.29).}
\end{subtable}
\vspace{-8pt}
\caption{\textbf{Ablation studies on image generation model designs.} 
% The evaluations are on 256$\times$256 ImageNet 50k validation set.
}
\vspace{-18pt}
\label{tab:token_design}
\end{table}

\vspace{2pt}
\noindent\textbf{Initial sequential token generation.} We first evaluate the importance of initial sequential token generation by comparing models with and without this phase in Tab.~\ref{tab:token_design} (a). Results show that initial sequential generation reduces FID from 3.67 to 2.61, with only 3 additional steps (147 vs. 144). We also visualize the comparison in Fig.~\ref{fig:exp_vis}. Without initial sequential generation (middle row), the generated images exhibit inconsistent global structures, such as misaligned dogs with duplicated body parts, as initial tokens are generated without awareness of each other. In contrast, our approach with initial sequential generation (top row) produces more coherent and natural-looking images. The results illustrate the importance of initial sequential token generation for establishing proper global structure.

\noindent\textbf{Number of parallel predicted tokens.} 
The number of tokens predicted in parallel ($n$) controls the trade-off between efficiency and quality.  As shown in Tab.~\ref{tab:token_design}(b), with $n=4$ ($M=2$), our approach reduces generation steps from 576 to 147 while maintaining comparable quality (FID 2.35 vs. 2.34). Further increasing to $n=16$ ($M=4$) achieves more aggressive parallelization with only 51 steps, at the cost of slight quality degradation (FID increase of 0.67). This is consistent with our analysis that tokens from distant regions have weaker dependencies and can be generated in parallel. As shown in Fig.~\ref{fig:intro_vis}, both PAR-4× and PAR-16× preserve visual fidelity while achieving significant speedup (3.46s and 1.31s vs. 12.41s).

% \textcolor{red}{[JS: add some motivation here. Otherwise cannot understand why we need to care about it. like the group size trades off the generation efficiency and quality.]} We study how the number of tokens predicted in parallel affects the generation quality and efficiency. This number corresponds to the size of region partition ($M\times M$), where $M\times M$ tokens are predicted simultaneously. We experiment with $M=1, 2, 4$ (i.e., group sizes of $n=1, 4,  16$). When $M=1$, the model degenerates to traditional token-by-token generation, serving as our baseline. With $M=2$, our approach achieves 4$\times$ speedup by reducing generation steps from 576 to 147, while maintaining comparable quality (FID 2.35 vs.\ 2.34, IS 263.24 vs.\ 253.90). Further increasing to $M=4$ provides larger speed improvement with 11.3$\times$ fewer steps (51 vs.\ 576), with only minor quality degradation (FID increase of 0.67). \textcolor{red}{[JS: the following conclusion is too straightforward and not attractive. Do you have result visualization of different group size? or any interesting observations? And can say these results  support that weakly dependent tokens can reliably generate together without hurting performance. ]} These results demonstrate that our approach can effectively balance generation quality and efficiency through appropriate parallel group size selection.

\vspace{1pt}
\noindent\textbf{Impact of attention pattern.} 
To enable effective parallel prediction while preserving rich context modeling, we study different attention patterns between parallel predicted tokens. With $n=4$ parallel tokens, enabling full attention within groups reduces FID from 3.64 to 2.61 compared to causal attention, as it allows each token to access complete context from previous groups. This supports our design of combining bi-directional attention within groups with autoregressive attention between groups.
\vspace{1pt}
\noindent\textbf{Impact of token ordering and prediction pattern.}
We compare raster scan and our distant ordering under different prediction settings. As shown in Tab.~\ref{tab:token_design}(d), while both achieve comparable quality in single-token prediction (FID 2.62 vs. 2.64), their performance differs significantly with multi-token prediction - raster scan degrades severely (FID 5.64) while our distant ordering maintains quality (FID 2.61). This indicates that the choice of parallel predicted tokens is critical. When using raster scan, adjacent tokens with strong dependencies are forced to generate simultaneously, leading to distorted local patterns as shown in Fig.~\ref{fig:exp_vis} (bottom row). In contrast, our region-based distant ordering groups weakly correlated tokens for parallel prediction, preserving both local details and global coherence (top row).

\vspace{1pt}
\noindent\textbf{Model scaling analysis.} 
In Tab.~\ref{tab:token_design}(e), we study how our parallel prediction approach scales with model size. With $n=4$ parallel tokens, increasing model size from 343M to 3.1B parameters steadily improves generation quality (FID decreases from 3.76 to 2.29). Comparing with sequential generation baseline (LlamaGen) in Tab.~\ref{tab:main}, while smaller models show a noticeable quality gap (343M: FID 3.76 vs. 3.07), larger models achieve comparable performance (775M: 2.61 vs. 2.62; 1.4B: 2.35 vs. 2.34) while reducing generation steps from 576 to 147. This demonstrates that increased model capacity helps mitigate the quality trade-off from parallel prediction, suggesting stronger capability in modeling joint distribution of parallel tokens.
% \noindent\textbf{Model scaling analysis.} We investigate whether our parallel prediction approach scales well to larger models by evaluating four variants from 343M to 3.1B parameters in Tab.\ref{tab:token_design}(e). With $n=4$ parallel prediction, FID scores steadily improve from 3.76 to 2.29 as model size increases, while maintaining the same efficiency benefits (around 4× faster generation). This demonstrates the effectiveness of our approach across different model scales.

\section{Conclusion}
\label{sec:conclusion}

We propose PAR, a noval autoregressive visual generation approach that enables efficient parallel generation while preserving the advantages of autoregressive modeling. Our key finding is that the feasibility of parallel generation depends on token dependencies - tokens with weak dependencies can be generated in parallel while strongly dependent tokens lead to inconsistent results. Based on this insight, our \our organizes tokens based on their dependency strengths rather than spatial proximity. The effectiveness of our approach across different visual domains validates this strategy for efficient autoregressive visual generation. We hope our work can inspire future research on visual generation and other sequence prediction tasks.
\section*{Acknowledgements}
This work is partially supported by the National Nature Science Foundation of China (No. 62402406). The authors would like to thank Peize Sun, Yi Jiang, Xian Liu and Yao Teng for helpful discussions on autoregressive models.

{
    \small
    \bibliographystyle{ieeenat_fullname}
    \bibliography{main}
}

\clearpage
\setcounter{page}{1}
\setcounter{section}{0}
\renewcommand{\thesection}{\Alph{section}}
\maketitlesupplementary

\section*{Appendix}
\addcontentsline{toc}{section}{Appendix}

The supplementary material includes the following additional information:
\begin{itemize}
    \item Sec.~\ref{sec:A} provides more implementation details for PAR.
    \item Sec.~\ref{sec:B} demonstrates the compatibility of our approach with typical LLM engineering optimizations.
    \item Sec.~\ref{sec:C} provides more visualization results, including zero-shot high-resolution generation and long-range dependency examples.
    \item Sec.~\ref{sec:D} provides the analysis of visual token dependencies.
\end{itemize}

\section{Implementation details for PAR}
\label{sec:A}

\textbf{Image Generation.} For image generation, we train our models on the ImageNet-1K~\cite{imagenet} training set, consisting of 1,281,167 images across 1,000 object classes. Following the setting in LlamaGen~\cite{sun2024autoregressive}, we pre-tokenize the entire training set using their VQGAN~\cite{esser2020taming} tokenizer and enhance data diversity through ten-crop transformation. For inference, we adopt classifier-free guidance~\cite{ho2022classifier} to improve generation quality. The detailed training and sampling hyper-parameters are listed in Tab.~\ref{tab:img_params}.

\begin{table}[h]
\centering
\begin{tabular}{p{0.45\columnwidth}|p{0.45\columnwidth}}
% \toprule
config & value \\
\Xhline{1.2pt}
\multicolumn{2}{c}{\textit{training hyper-params}} \\
\Xhline{0.8pt}
optimizer & AdamW~\cite{loshchilov2017adamw} \\
learning rate & 1e-4(L,XL)/2e-4(XXL,3B)  \\
weight decay & 5e-2 \\
optimizer momentum & (0.9, 0.95) \\
batch size & 256(L,XL)/ 512(XXL,3B) \\
learning rate schedule & cosine decay \\
ending learning rate & 0 \\
total epochs & 300 \\
warmup epochs & 15 \\
precision & bfloat16 \\
max grad norm & 1.0 \\
dropout rate & 0.1 \\
attn dropout rate & 0.1 \\
class label dropout rate & 0.1 \\
\Xhline{0.8pt}
\multicolumn{2}{c}{\textit{sampling hyper-params}} \\
\Xhline{0.8pt}
temperature & 1.0\\
guidance scale & 1.60 (L) / 1.50 (XL) / 1.435 (XXL) / 1.345 (3B) \\
\Xhline{0.8pt}
\end{tabular}
\caption{\textbf{Detailed Hyper-parameters for Image Generation.}}
\label{tab:img_params}
\end{table}

\noindent\textbf{Video Generation.} For video generation, we train our models on the UCF-101~\cite{soomro2012ucf101} training set, which contains 9.5K training videos spanning 101 action categories. Videos are processed as 8fps random clips and tokenized by our reimplementation of MAGVIT-v2~\cite{yu2023language} (as their code is not publicly available), achieving a reconstruction FVD score of 32 on UCF-101. For inference, we use classifier-free guidance~\cite{ho2022classifier} with top-k sampling to improve generation quality. The detailed training and sampling hyper-parameters are listed in Tab.~\ref{tab:video_params}.

\begin{table}[h]
\centering
\begin{tabular}{p{0.45\columnwidth}|p{0.45\columnwidth}}
config & value \\
\Xhline{1.2pt} 
\multicolumn{2}{c}{\textit{training hyper-params}} \\
\Xhline{0.8pt}
optimizer & AdamW~\cite{loshchilov2017adamw} \\
learning rate & 1e-4 \\
weight decay & 5e-2 \\
optimizer momentum & (0.9, 0.95) \\
batch size & 256 \\
learning rate schedule & cosine decay \\
ending learning rate & 0 \\
total epochs & 3000 \\
warmup epochs & 150 \\
precision & bfloat16 \\
max grad norm & 1.0 \\
dropout rate & 0.1 \\
attn dropout rate & 0.1 \\
class label dropout rate & 0.1 \\
\Xhline{0.8pt}
\multicolumn{2}{c}{\textit{sampling hyper-params}} \\
\Xhline{0.8pt}
temperature & 1.0\\
guidance scale & 1.15 \\
top-k & 8000 \\
\Xhline{0.8pt}
\end{tabular}
\caption{\textbf{Detailed Hyper-parameters for Video Generation.}}
\label{tab:video_params}
\end{table}

\section{Compatibility with Typical LLM Engineering Optimizations}
\label{sec:B}

We investigate whether our algorithmic parallel generation approach can complement typical engineering optimizations used in LLM inference. All experiments were conducted on a single NVIDIA A100 GPU with batch size 1, generating 384×384 resolution images.
For simplicity, we only implemented PyTorch's compile feature (leveraging CUDA graph optimization) in our PAR model. As a comparison point, we tested LlamaGen~\cite{sun2024autoregressive} with vLLM~\cite{kwon2023efficient} optimizations, which includes both page attention and CUDA graph optimizations.

\begin{table}[h]
\centering
\begin{tabular}{l|c|c|c}
\toprule
Model & Resolution & Optimization & Latency \\
\midrule
LlamaGen-3B &  384 & none & 12.41s \\
LlamaGen-3B &  384 & vLLM & 4.12s \\
PAR-3B-4x & 384 & none & 3.46s \\
PAR-3B-4x &  384 & compile & 1.15s \\
PAR-3B-16x &  384 & compile & 0.43s \\
\bottomrule
\end{tabular}
\vspace{-6pt}
\caption{\textbf{Compatibility with LLM engineering optimizations.} Even with just PyTorch compile optimization, our PAR approach achieves substantial speedups compared to LlamaGen with more comprehensive vLLM optimizations.}
\label{tab:eng_speedup}
\end{table}

As shown in Tab.~\ref{tab:eng_speedup}, our algorithmic improvements and engineering optimizations are orthogonal and provide complementary benefits. Even without any engineering optimization, PAR-3B-4x (3.46s) outperforms LlamaGen-3B with comprehensive vLLM optimizations (4.12s). When implementing just the simple CUDA graph optimization through PyTorch compile, PAR-3B-4x achieves 1.15s latency, a 3.6× improvement over optimized LlamaGen.
With more aggressive parallelization, PAR-3B-16x with compile further reduces latency to 0.43s, demonstrating our approach's flexibility in speed-quality trade-offs. These results confirm that algorithm-level optimizations (reducing sequential steps) and engineering-level optimizations (improving computational efficiency) are orthogonal approaches that, when combined, maximize generation efficiency beyond what either can achieve alone.

\section{More Visualization Results}
\label{sec:C}

\begin{figure}[!t]
  \centering
  \includegraphics[width=\linewidth]{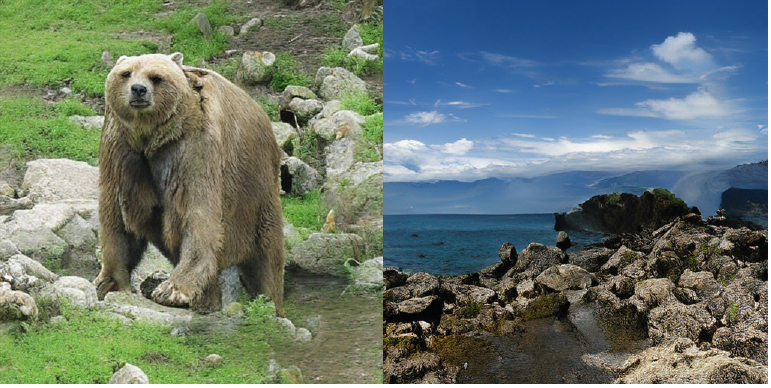}
  \vspace{-20pt}
  \caption{\textbf{Zero-shot generation at 512×512 resolution.} Our model successfully generates coherent high-resolution images despite being trained at 384×384 resolution.}
  \vspace{-6pt}
  \label{fig:high_res}
\end{figure}

\noindent\textbf{Zero-shot Generation on Higher Resolutions}.
Fig.~\ref{fig:high_res} demonstrates our model's capability for zero-shot generation at higher resolutions (512×512) using Rotary Position Embedding~\cite{su2024roformer}. Despite being trained on 384×384 images, our approach effectively maintains coherent global structures and detailed local patterns in higher resolution generation. This shows the flexibility of our parallel generation framework and its compatibility with positional encoding methods that support resolution extrapolation.

\begin{figure}[h]
  \centering
  \includegraphics[width=\linewidth]{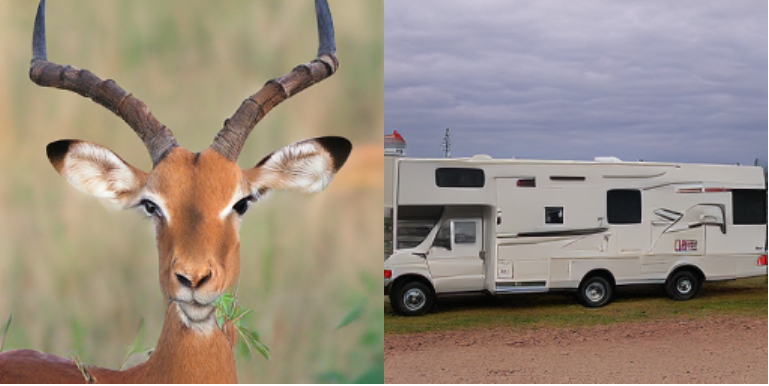}
  \vspace{-20pt}
  \caption{\textbf{Long-range dependency handling.} Our method successfully maintains consistency between distant but strongly related elements (deer antlers, vehicle wheels), even when generating tokens from different spatial regions in parallel.}
  \vspace{-6pt}
  \label{fig:long_range}
\end{figure}

\noindent\textbf{Long-range Dependency Handling}
While our approach leverages the observation that spatially distant tokens typically have weaker dependencies, certain visual elements exhibit strong long-range dependencies. Fig.~\ref{fig:long_range} showcases our model's ability to maintain consistency between distant but strongly dependent visual elements, such as symmetric features (deer antlers, vehicle wheels) and coherent structures across the image.

\begin{figure*}[tbp]
  \centering
  \includegraphics[width=\linewidth]{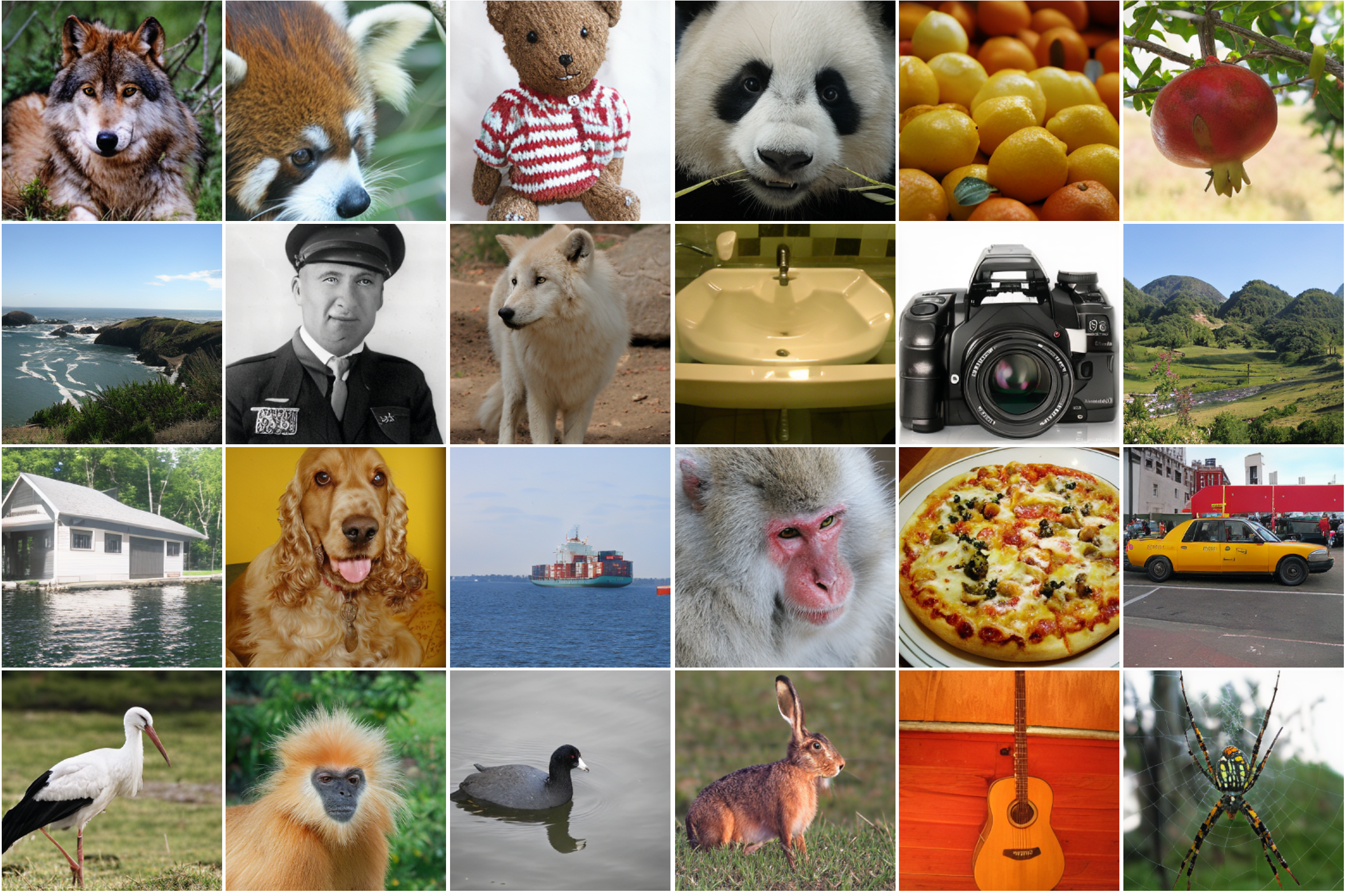}
  \vspace{-16pt}
  \caption{\textbf{Additional image generation results of PAR-4$\times$ across different ImageNet~\cite{imagenet} categories.}
  }
  \label{fig:supp_vis_4}
   \vspace{-4pt}
\end{figure*}

\begin{figure*}[h]
  \centering
  \includegraphics[width=\linewidth]{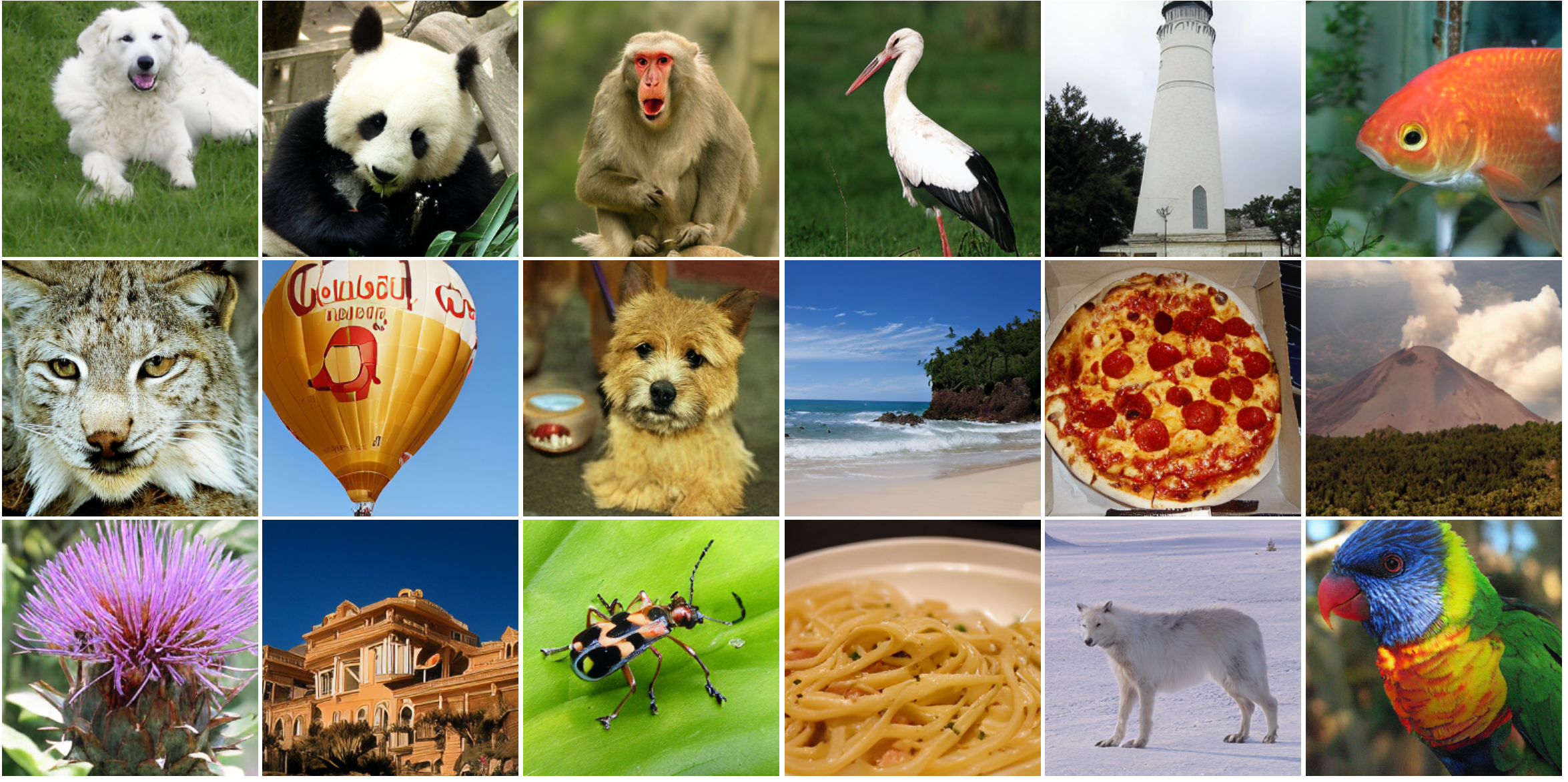}
  \vspace{-16pt}
  \caption{\textbf{Additional image generation results of PAR-16$\times$ across different ImageNet~\cite{imagenet} categories.}
  }
  \label{fig:supp_vis_16}
   \vspace{-16pt}
\end{figure*}

\noindent\textbf{Addtional Image Generation Visualization}. In Fig.\ref{fig:supp_vis_4} and Fig.\ref{fig:supp_vis_16}, we provide additional visualization results of PAR-4$\times$ and PAR-16$\times$ image generation on ImageNet~\cite{imagenet} dataset, respectively.

\begin{figure*}[h]
  \centering
  \includegraphics[width=\linewidth]{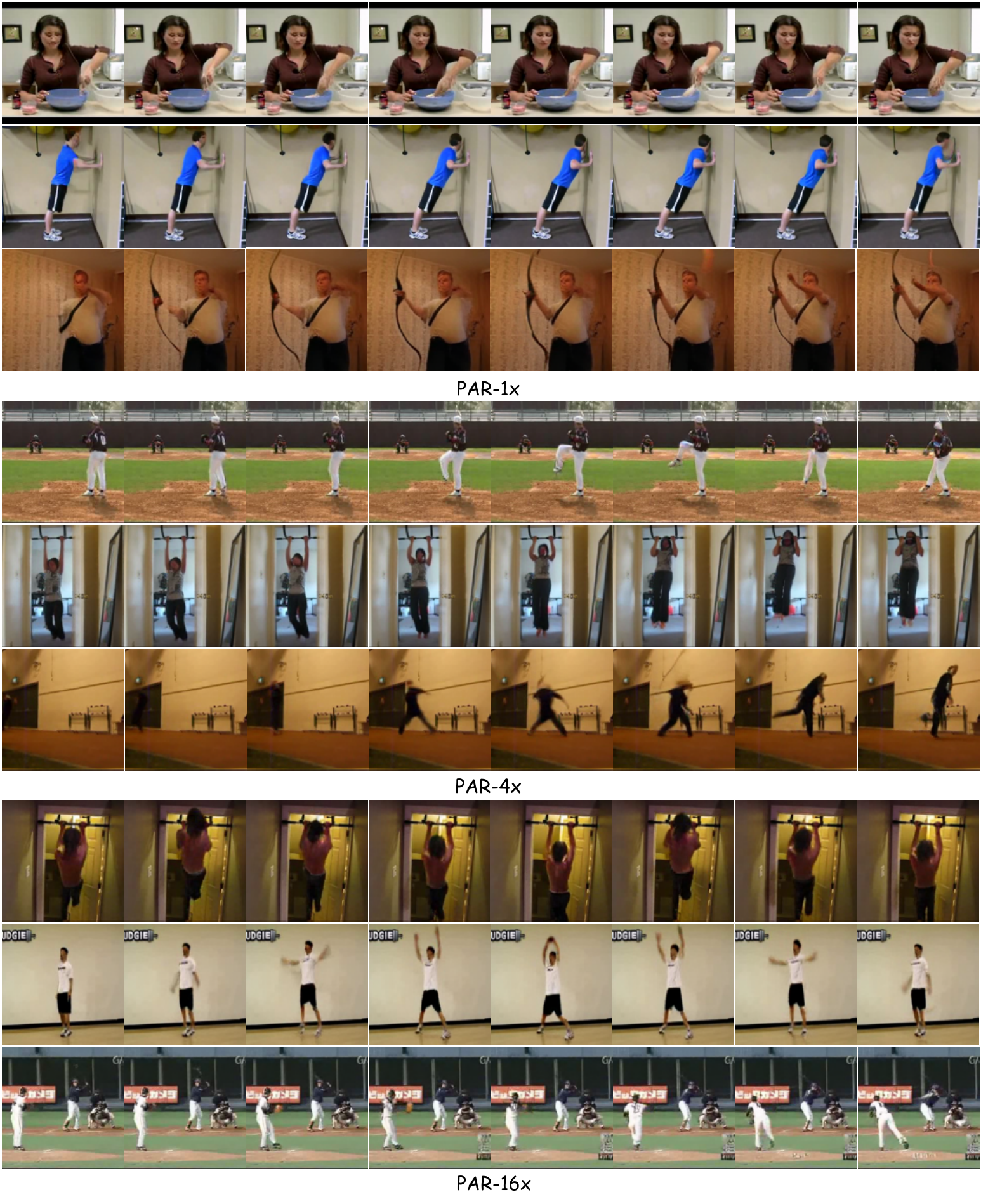}
  \vspace{-16pt}
  \caption{\textbf{Video generation results on UCF-101~\cite{soomro2012ucf101}.} Each row shows sampled frames from a 17-frame sequence at 128×128 resolution, generated by PAR-1$\times$, PAR-4$\times$, and PAR-16$\times$ respectively across different action categories.}
  \label{fig:supp_video}
   \vspace{-16pt}
\end{figure*}

\noindent\textbf{Addtional Video Generation Visualization}. In Fig.\ref{fig:supp_video}, we provide the visualization results of video generation using our model on the UCF-101\cite{soomro2012ucf101} dataset. The results are sampled from 128×128 resolution videos with 17 frames. As shown in the figure, even with 16× parallelization (PAR-16$\times$), our method shows no obvious quality degradation compared to single-token prediction (PAR-1$\times$), producing smooth motion and stable backgrounds across frames.

\section{Analysis of Visual Token Dependencies}
\label{sec:D}
In Sec.\ref{sec:3.1}, we demonstrated through pilot studies that parallel generation of adjacent tokens leads to quality degradation due to strong dependencies, while tokens from distant regions can be generated simultaneously. In this section, we provide a theoretical perspective of conditional entropy to explain this observation and our design. We use conditional entropy to measure the token dependencies quantitatively - lower conditional entropy between tokens indicates stronger dependency, while higher conditional entropy suggests weaker dependency and thus potential for parallel generation. We further validate our PAR design from the perspective of conditional entropy - In AR-based generation, each step predicts a conditional distribution of the next tokens given all previous tokens. Higher conditional entropy indicates higher difficulty for the model to predict the next tokens. In this section, we first introduce the estimation of conditional entropy in Sec.\ref{subsec:entropy-estimation}, and then validate our proposed approach by analyzing the relationship between token dependencies and spatial distances in Sec.~\ref{subsec:entropy-analysis}.

\subsection{Conditional Entropy Estimation}\label{subsec:entropy-estimation}
Given a visual token sequence $\{v_1, v_2, ..., v_n\}$, our goal is to estimate the conditional entropy $H(v_{k}|\{v_j\}_{j < k})$  where the token feature $v_i \in \mathbb{R}^d$ and $\{v_j\}_{j < k}$ is the set of (all) visual tokens that precede $v_k$ in the sequence. This conditional entropy measures the uncertainty of the current token $v_k$ given the previously occurring visual tokens, thereby characterizing the dependency between $v_k$ and the set $\{v_j\}_{j < k}$. It is important to emphasize that we do not require the exact value of $H(v_{k}|\{v_j\}_{j < k})$. Instead, we aim to reflect the trends in $H(v_{k}|\{v_j\}_{j < k})$ under different scenarios, such as given different sets of $\{v_j\}_{j < k}$ and  considering different positions of $v_k$ given the same set of $\{v_j\}_{j < k}$.

In particular, we characterize the relationship between the token $v_k$ and the previous ones as the following model
\begin{equation}
\label{eq:model}
v_{k} = f(\{v_j\}_{j < k}) + \boldsymbol{\epsilon}_k
\end{equation}
where $v_{k}$ is the next token we focus on and $\{v_j\}_{j < k}$ is the conditioning token(s), $f(\cdot)$ is a deterministic function, and $\boldsymbol{\epsilon}_k$ is the random additive error term. Then the conditional entropy $H(v_{k}|\{v_j\}_{j < k})$ satisfies
\begin{align}
H(v_{k}|\{v_j\}_{j < k}) &= H(f(\{v_j\}_{j < k}) + \boldsymbol{\epsilon}_k|\{v_j\}_{j < k}) \notag \\
&=H(\boldsymbol{\epsilon}_k|\{v_j\}_{j < k}),
\end{align}
where the second equation holds since $f(\cdot)$ is a deterministic function. However, exactly calculating $H(\boldsymbol{\epsilon}_k|\{v_j\}_{j < k})$ is intractable as we cannot access the entire data distribution. To this end, inspired by prior research on bounding techniques for entropy and mutual information estimation \cite{belghazi2018mine,nguyen2010estimating,oord2018representation,alemi2016deep,verdu2015alpha,tishby2015deep}, we seek their upper bound as a proxy for showing the trends of the conditional entropy for different tokens. In particular, we have
\begin{align}\label{eq:closed-form}
H(\boldsymbol{\epsilon}_k|\{v_j\}_{j < k})\le H(\boldsymbol{\epsilon}_k)\le\frac{1}{2}\log((2\pi e)^d|\boldsymbol{\Sigma}|),
\end{align}
where $\boldsymbol{\Sigma}$ denotes the covariance matrix of $\boldsymbol{\epsilon}_k$. Notably, the first inequality naturally holds and the second inequality follows from the maximum entropy theory \cite{jaynes2003probability,cover1999elements}, which is achievable when $\boldsymbol{\epsilon}_k$ follows a Gaussian distribution.

Based on Eq.~\ref{eq:closed-form}, we can estimate the trend of conditional entropy changes by calculating the determinant of the residual covariance matrix, i.e., $|\boldsymbol{\Sigma}|$. In order to obtain the additive errors $\boldsymbol{\epsilon}$, we consider training a parameterized model $f_\theta(\cdot)$ to get the function $f$ and characterize $\boldsymbol{\epsilon}$ as the residual errors. The detailed algorithm is provided in Algorithm \ref{alg:conditional_entropy_estimation}.

\begin{figure}[htbp]
\centering
\begin{subfigure}[b]{0.24\linewidth}
\includegraphics[width=\linewidth]{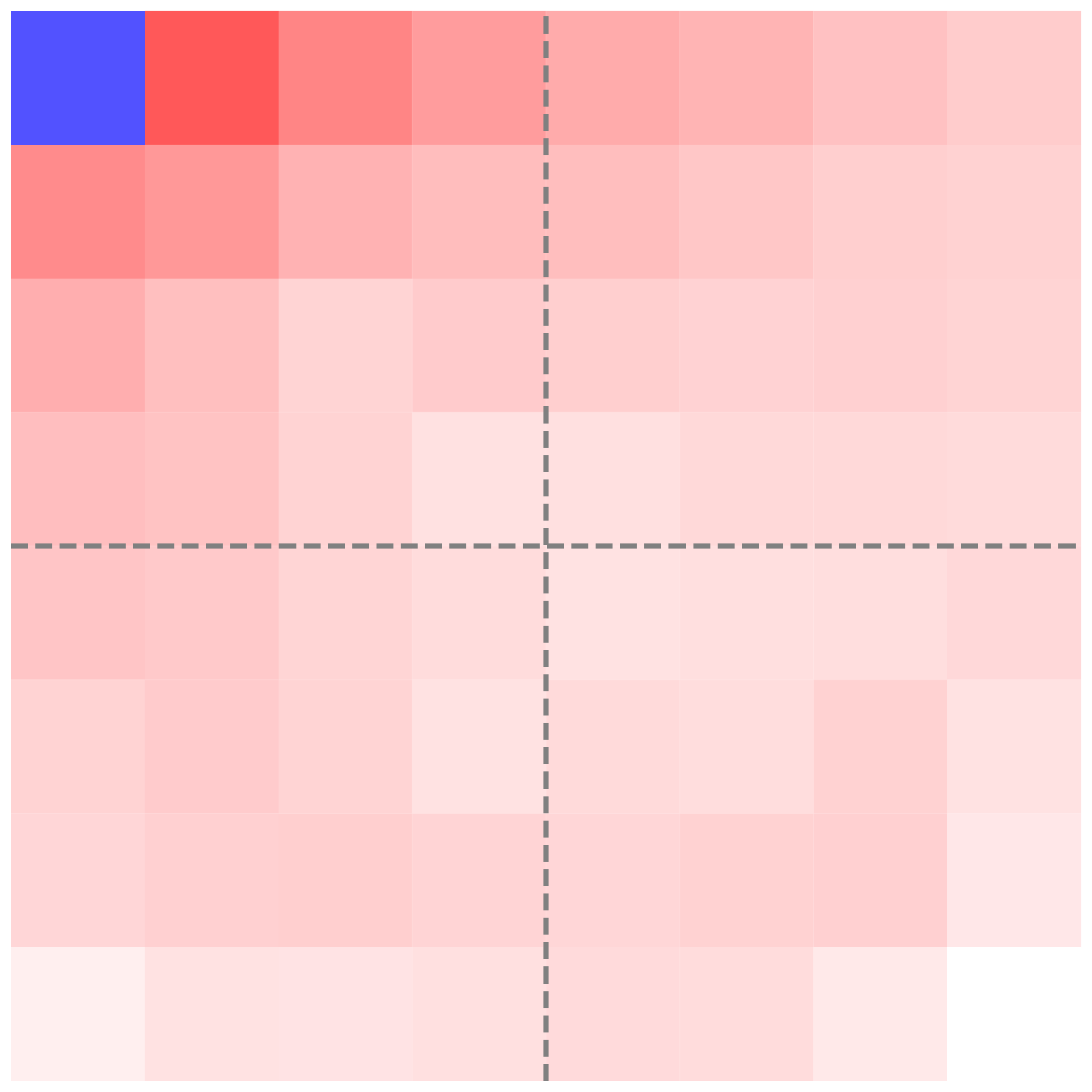}
\end{subfigure}
\hfill
\begin{subfigure}[b]{0.24\linewidth}
\includegraphics[width=\linewidth]{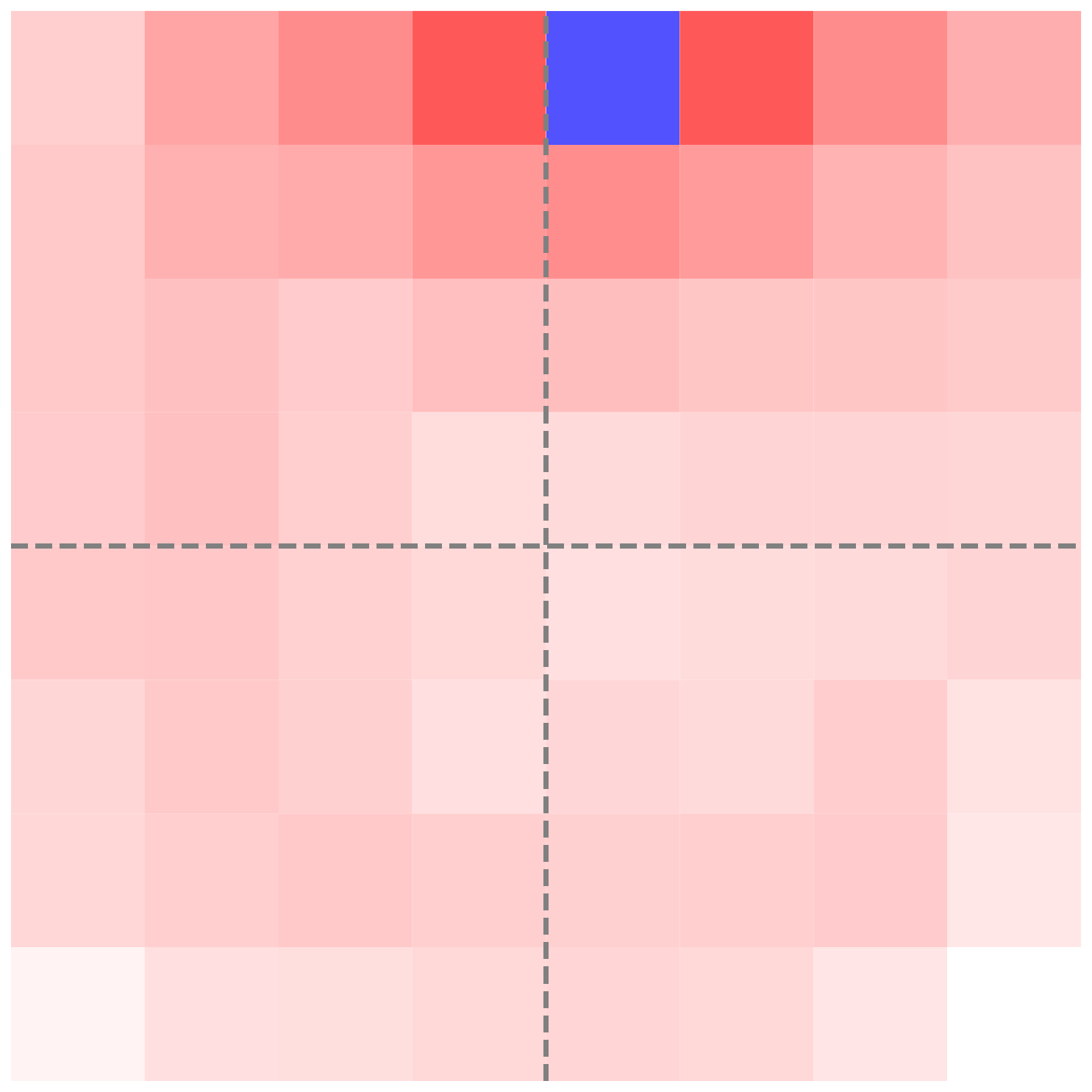}
\end{subfigure}
\hfill
\begin{subfigure}[b]{0.24\linewidth}
\includegraphics[width=\linewidth]{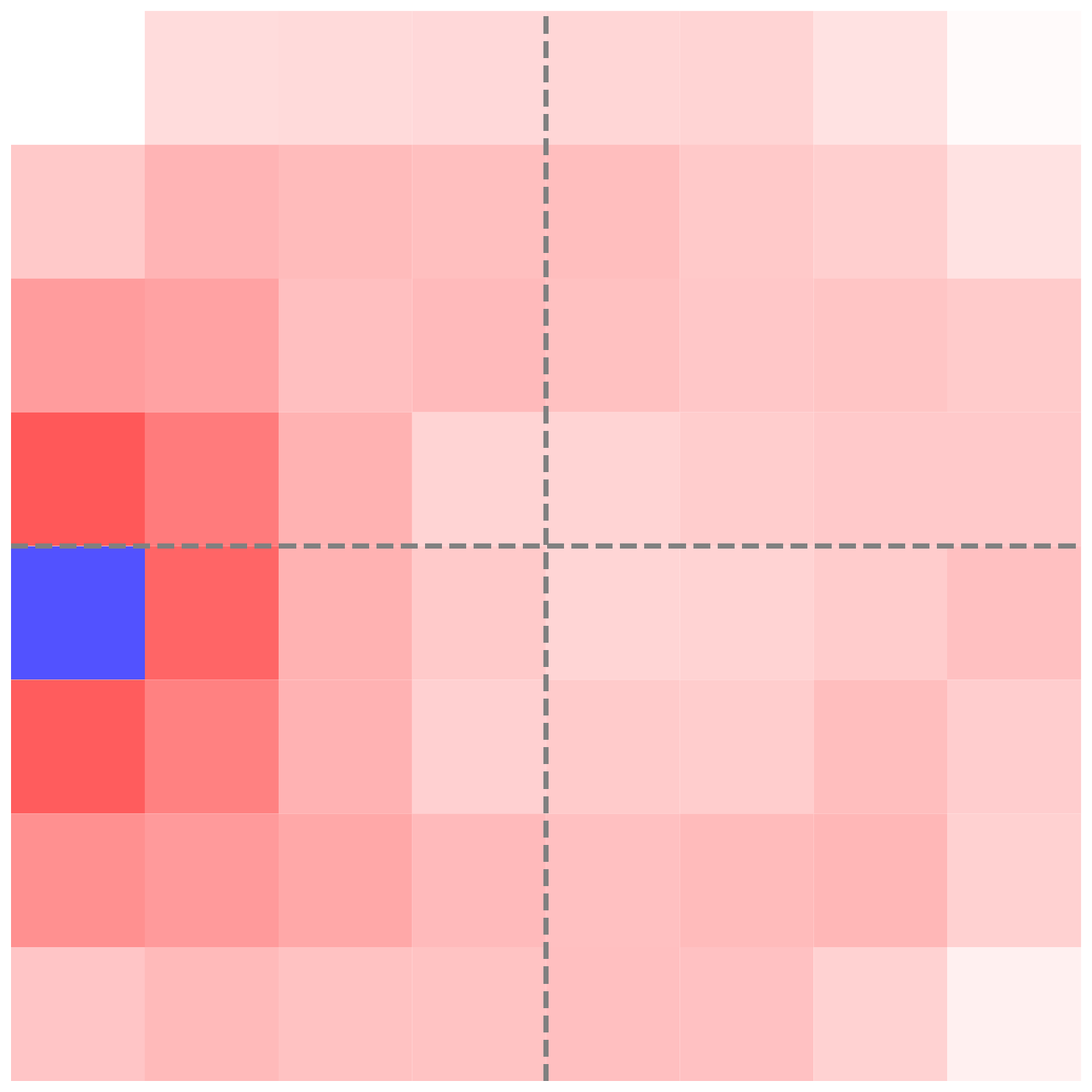}
\end{subfigure}
\hfill
\begin{subfigure}[b]{0.24\linewidth}
\includegraphics[width=\linewidth]{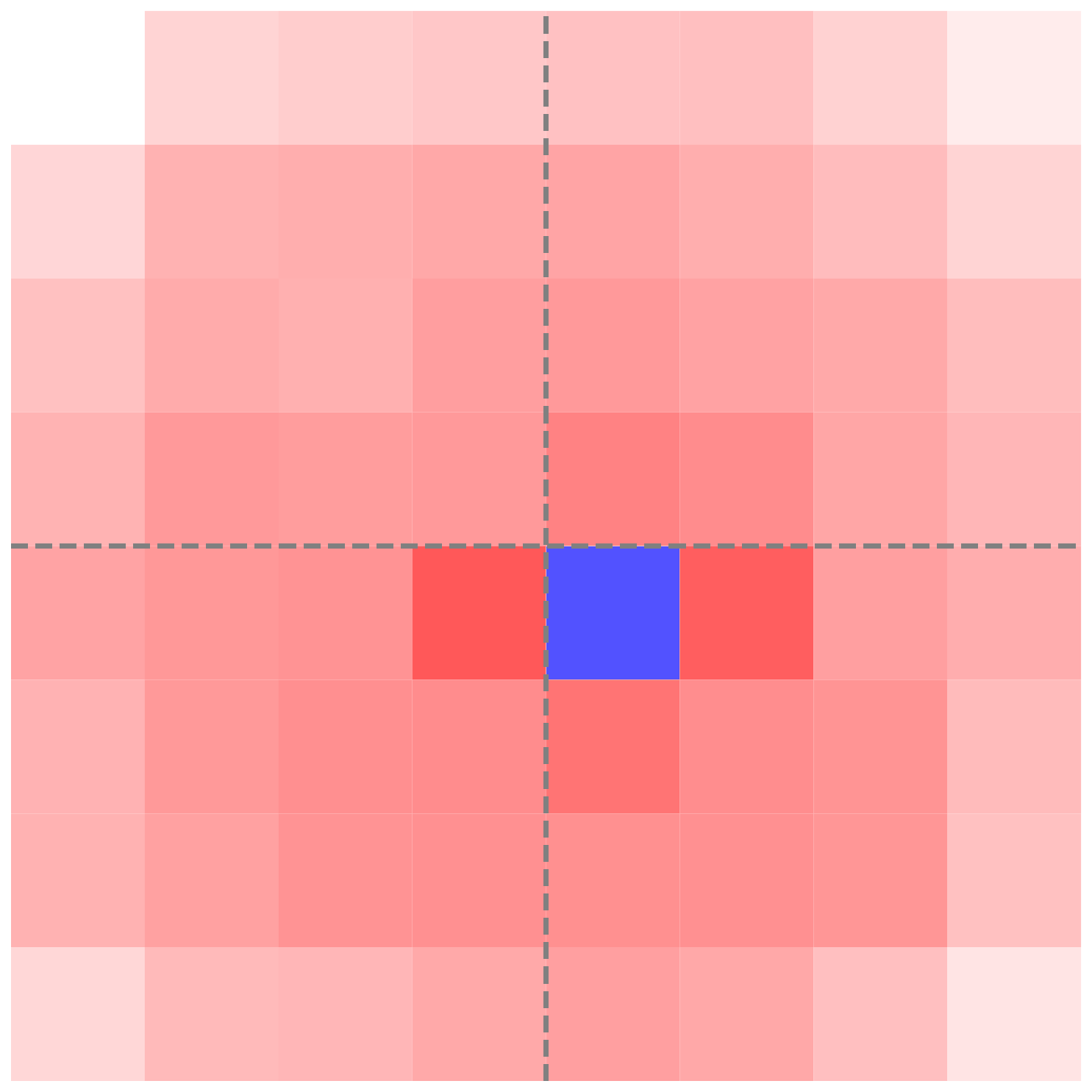}
\end{subfigure}
\vspace{-5pt}
\caption{Visualization of token conditional entropy maps. Each map shows the conditional entropy of all tokens when conditioned on a reference token (blue square). Darker red indicates lower conditional entropy and thus stronger dependency with the reference token. The visualization shows that \textbf{tokens exhibit strong dependencies with their spatial neighbors and weak dependencies with distant regions.}}
\label{fig:entropy}
% \vspace{-10pt}
\end{figure}

\begin{algorithm}
\caption{Conditional Entropy Estimation}
\begin{algorithmic}[1]
\Require 
    \State $m$: number of data points
    \State $\{v_{i,1}, v_{i,2}, ..., v_{i,n}\}_{i=1}^m$: visual token sequences, where each $v_{i,j} \in \mathbb{R}^d$
    \State $k$: index of the target token
    \State $f_{\theta}$: parameterized model
\Ensure Estimated conditional entropy $\hat{H}(v_{k}|\{v_j\}_{j < k})$
\State Initialize empty lists $\mathcal{X}$ and $\mathcal{Y}$
\For{$i = 1$ to $m$}
    \State $X_i \gets \{v_{i,j}\}_{j < k}$
    \State $Y_i \gets v_{i,k}$
    \State Append $(X_i, Y_i)$ to $(\mathcal{X}, \mathcal{Y})$
\EndFor
\State Train a model $f_{\theta}$ to estimate $Y$ given $X$ using $(\mathcal{X}, \mathcal{Y})$
\State Initialize empty list $\mathcal{E}$ for residuals
\For{$(X, Y)$ in $(\mathcal{X}, \mathcal{Y})$}
    \State $Y_{pred} \gets f_{\theta}(X)$
    \State $\boldsymbol{\epsilon}_k \gets Y - Y_{pred}$
    \State Append $\boldsymbol{\epsilon}_k$ to $\mathcal{E}_k$
\EndFor
\State Compute residual covariance matrix $\boldsymbol{\hat{\Sigma}}$ of $\mathcal{E}_k$
\State $\hat{H}(v_{k}|\{v_j\}_{j < k}) \gets \frac{1}{2}\log((2\pi e)^d|\boldsymbol{\hat{\Sigma}}|)$
\State \Return $\hat{H}(v_{k}|\{v_j\}_{j < k})$
\end{algorithmic}
\label{alg:conditional_entropy_estimation}
\end{algorithm}

\begin{figure}[h]
\centering
\includegraphics[width=\linewidth]{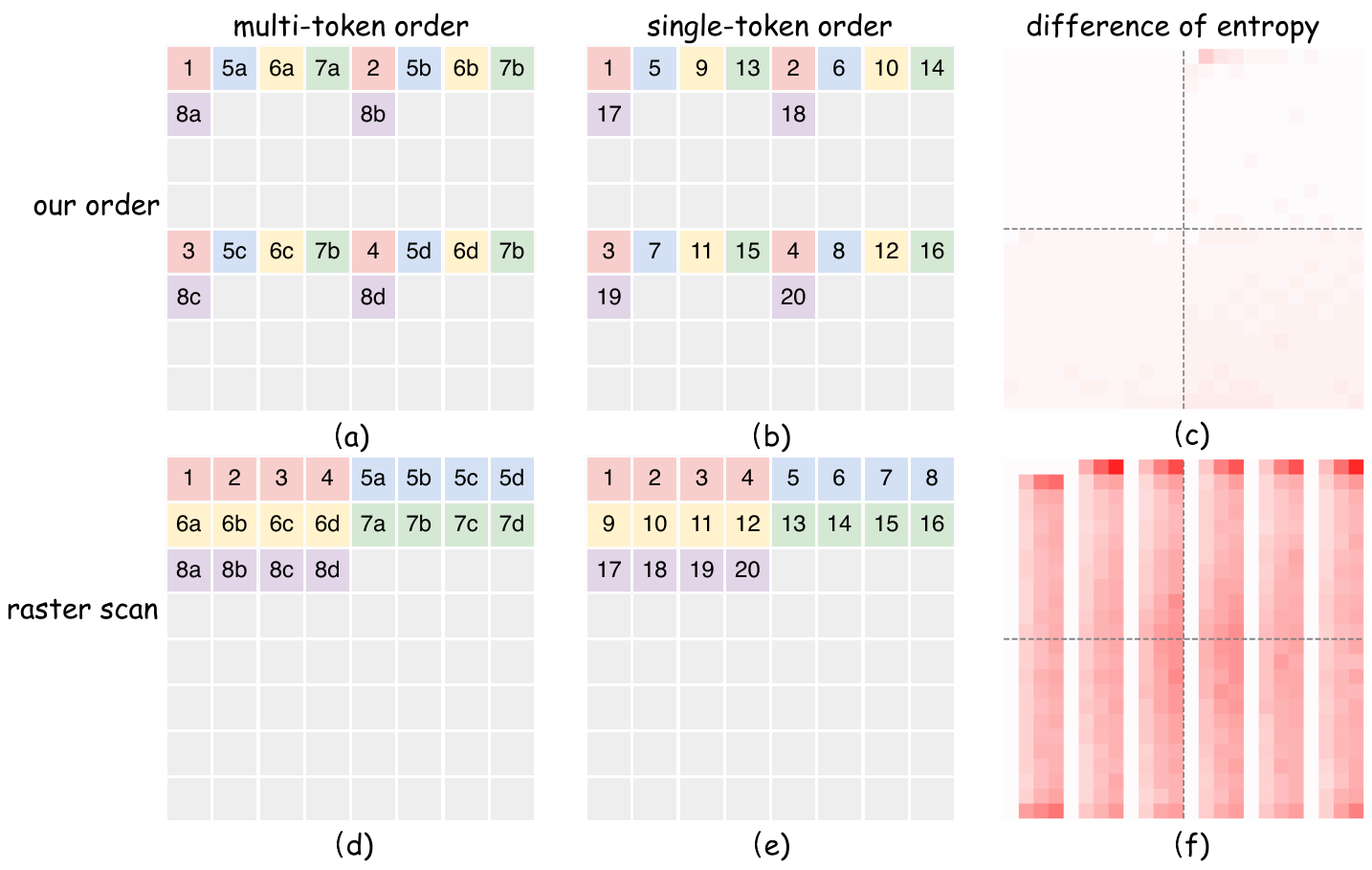}
\hfill
\vspace{-10pt}

\caption{\textbf{Conditional entropy differences between parallel and sequential generation in different orders.}
(a)(d) show parallel (4 tokens) generation strategies and (b)(e) show sequential generation strategies for our proposed order and raster scan order respectively. Numbers indicate generation step in each order. (c)(f) visualize the conditional entropy increase when switching from sequential to parallel generation for each order, where darker red indicates larger entropy increase and thus higher prediction difficulty. Both orders generate the first four tokens sequentially (shown as white regions in entropy maps). Our proposed order that generates tokens from different spatial blocks in parallel shows smaller entropy increases compared to raster scan order that generates consecutive tokens simultaneously, indicating parallel generation across spatial blocks introduces less prediction difficulty than generating adjacent tokens simultaneously.
}
\label{fig:entropy_diff}
\vspace{-10pt}
\end{figure}

\subsection{Entropy Analysis on ImageNet Data and PAR}\label{subsec:entropy-analysis}

Based on the conditional entropy estimation method introduced above, we conduct experiments on ImageNet to analyze token dependencies and validate our parallel generation strategy. We randomly sample 10,000 images from ImageNet~\cite{imagenet} and extract their features using VQGAN~\cite{esser2020taming} encoder, followed by vector quantization to obtain continuous features from the codebook. 

We first analyze the dependencies between tokens at different positions. 
For each position $j$ in the feature map, we calculate the conditional entropy $H(v_i|v_j)$ where $i \neq j$, given the token $v_j$ at the $j$-th position and considering all tokens $v_i$ at other positions.  It should be noted that Algorithm \ref{alg:conditional_entropy_estimation} is not limited to $H(v_{k}|\{v_j\}_{j < k})$ where the given visual tokens $\{v_j\}$ must satisfy $j<k$. This is because any given tokens $v_j$ and $v_i$ can be considered to satisfy Eq.~\ref{eq:model}, making the proposed method applicable in calculating $H(v_i|v_j)$. Fig.~\ref{fig:entropy} presents the experimental results. We observe that given different token positions $v_i$, the adjacent tokens typically exhibit lower conditional entropy (shown in redder colors). This indicates that the dependencies between adjacent tokens are stronger compared to the dependencies between tokens that are farther apart in position. 
This observation aligns with the spatial locality in visual data, where nearby regions have stronger correlations than distant ones.

Next, we analyze how different token ordering strategies affect the difficulty of parallel generation in Fig.~\ref{fig:entropy_diff}. To simulate the prediction difficulty during generation, we compute each token\textquotesingle s conditional entropy given all its previous tokens - higher conditional entropy indicates more uncertainty and thus higher prediction difficulty at that position.  By comparing the conditional entropy difference between sequential (one token at a time) and parallel generation (predicting multiple tokens simultaneously), we can quantify the increased difficulty introduced by parallel generation at each position. 
 We conduct experiments with 4-token parallel prediction under two ordering strategies: our proposed generation order that first generates the initial four tokens sequentially to establish global structure, then generates tokens from different spatial blocks in parallel, and the raster scan ordering that directly predicts consecutive tokens simultaneously after the initial four tokens.
% Following notations in Sec.~\ref{sec:3.2}, the conditional entropy for these scenarios are defined as follows: 

For our proposed order, we aim to characterize the entropy increase caused by the parallel generation, when compared to the entirely sequential generation methods. In particular, let $v_k^{(r)}$ be the token at position $k$ in region $r$, we define $\mathcal V_{k,r}^{\mathrm{seq}}$ and $\mathcal V_{k,r}^{\mathrm{par}}$ by the sets of the previous tokens of $v_{k}^{(r)}$ for sequential and parallel generations (see Fig.~\ref{fig:entropy_diff}(a)(b)). Then the conditional entropy of the sequential generation (single-token) and parallel generation (multi-token) are defined as $H(v_k^{(r)}|\mathcal V_{k,r}^{\mathrm{seq}})$ and $H(v_k^{(r)}|\mathcal V_{k,r}^{\mathrm{par}})$. 
% Clearly, as $\mathcal V_{k,r}^{\mathrm{par}}\subseteq \mathcal V_{k,r}^{\mathrm{seq}}$,
We characterize the entropy increase caused by the parallel generation, i.e.,
\begin{align}\label{eq:diff_entropy_ourorder}
H(v_k^{(r)}|\mathcal V_{k,r}^{\mathrm{par}})-H(v_k^{(r)}|\mathcal V_{k,r}^{\mathrm{seq}}).
\end{align}

As a comparison, we also consider the raster scan order, where the tokens are exactly arranged based on their positions, denoted as $v_1,v_2,\ldots$. In this setting, given the current token $v_k$, we define $\mathcal V_{k}^{\mathrm{seq}}$ and $\mathcal V_{k}^{\mathrm{par}}$ by the sets of the previous tokens of $v_{k}$ for sequential and parallel generations (see Fig.~\ref{fig:entropy_diff}(d)(e)).
Then, we will also characterize the entropy increase caused by the parallel generation in the raster scan order, i.e.,
\begin{align}\label{eq:diff_entropy_raster}
H(v_k|\mathcal V_{k}^{\mathrm{par}})-H(v_k|\mathcal V_{k}^{\mathrm{seq}}).
\end{align}
The numerical results of \eqref{eq:diff_entropy_ourorder} and \eqref{eq:diff_entropy_raster} are presented in Fig.~\ref{fig:entropy_diff}(c) and (f). It can be seen that both orderings maintain identical conditional entropy for the first four tokens due to the sequential generation. For subsequent tokens, our proposed order leads to significantly smaller conditional entropy increases compared to the raster scan order. This indicates that when switching from sequential to parallel generation, generating tokens from different spatial blocks introduces less prediction difficulty than generating consecutive tokens in raster scan order. The result quantitatively validates our design.

\end{document}